\documentclass{article}
\usepackage{PRIMEarxiv}

\usepackage[english]{babel}

\usepackage{amsmath}
\usepackage{amsfonts}
\usepackage{listings}
\usepackage{graphicx}
\usepackage{parskip}
\usepackage{tabularx}
\usepackage{float}
\usepackage{geometry}
\usepackage{xcolor}
\usepackage{hyperref}
\usepackage{booktabs}
\usepackage{mathtools}
\usepackage{adjustbox}
\usepackage{makecell}
\usepackage{siunitx}
\usepackage{subcaption}
\usepackage[utf8]{inputenc} 
\usepackage[T1]{fontenc}    
\usepackage{url}            
\usepackage{nicefrac}       
\usepackage{microtype}      
\usepackage{lipsum}
\usepackage{fancyhdr}       
\graphicspath{{media/}}     
\usepackage{comment}
\usepackage{listings}
\usepackage{cleveref}
\usepackage{textcomp}

\usepackage[most]{tcolorbox}

\usetikzlibrary{arrows.meta, calc, backgrounds, positioning}
\usepackage{tikz}
\usepackage{tikz-3dplot}
\usetikzlibrary{shapes,arrows,calc,positioning,shadings}

\definecolor{codegreen}{rgb}{0,0.6,0}
\definecolor{codegray}{rgb}{0.5,0.5,0.5}
\definecolor{codepurple}{rgb}{0.58,0,0.82}
\definecolor{backcolour}{rgb}{0.95,0.95,0.92}

\PassOptionsToPackage{linktocpage}{hyperref}

\usepackage{amssymb}
\usepackage{pifont}

\definecolor{ForestGreen}{RGB}{34,139,34}
\definecolor{MidnightBlue}{RGB}{25, 25, 112}

\usepackage{tcolorbox}

\usepackage{tikz}
\usetikzlibrary{shapes,arrows,positioning}

\usepackage[colorinlistoftodos]{todonotes}

\hypersetup{
    colorlinks=true,
    urlcolor=MidnightBlue,
    linkcolor=ForestGreen,
    citecolor=ForestGreen
}

\definecolor{bg}{gray}{0.95}

\usepackage[
backend=biber,
style=authoryear,
]{biblatex}

\usepackage{xcolor}

\usepackage{csquotes}  
 \newcommand\blfootnote[1]{%
  \begingroup
  \renewcommand\thefootnote{}\footnote{#1}%
  \addtocounter{footnote}{-1}%
  \endgroup
}

\usepackage{scalerel}
\usepackage{stackengine,wasysym}

\def\sym#1{\ifmmode^{#1}\else\(^{#1}\)\fi}

\title{A Rosetta Stone for AI Benchmarks}

\author{
Anson Ho$^{\dagger}$, 
Jean-Stanislas Denain$^{\dagger}$, 
David Atanasov$^{\dagger}$,
Samuel Albanie$^{\ddagger}$,
Rohin Shah$^{\ddagger}$
}

\begin{document}

\maketitle

\vspace{-0.5cm}
\begin{abstract}
Most AI benchmarks saturate within years or even months after they are introduced, making it hard to study long-run trends in AI capabilities. To address this challenge, we build a statistical framework that stitches benchmarks together, putting model capabilities and benchmark difficulties on a single numerical scale. This acts as a ``Rosetta Stone", allowing us to compare models across a wide range of abilities and time, even if they are not evaluated on the same benchmarks. Moreover, this works without assuming how capabilities evolve across time or with training compute. We demonstrate three applications of this framework. First, we use it to measure the speed of AI progress over time, and to forecast future AI capabilities. Second, we estimate the rate of improvements in algorithmic efficiency, finding estimates that are higher, but broadly consistent with prior work. Finally, we find that our approach can be used to detect rapid accelerations in AI progress.
\end{abstract}

\blfootnote{$^\dagger$Epoch AI. $^\ddagger$Google DeepMind. This work was financially supported by Google DeepMind. Correspondence to \url{anson@epoch.ai}. All code can be found at: \url{https://github.com/epoch-research/benchmark-stitching}.}

\section{Introduction}
\label{sec:introduction}
Researchers typically use benchmarks to track AI progress, but each benchmark only gives a narrow picture of AI capabilities. Individual benchmarks saturate quickly, making it hard to study trends over long time periods. Moreover, we can only compare two models if they are evaluated on the same benchmark, which is often not the case in practice. 

In this paper, we introduce a simple statistical framework that addresses these challenges. By analyzing a wide collection of existing results, our method ``stitches'' these disparate benchmarks together, translating their scores onto a single, quantitative scale. This thus acts as a ``Rosetta Stone" for different benchmarks, allowing us to compare models evaluated on different benchmarks, and across long periods of time. 

\begin{figure}[H]
    \centering
    \includegraphics[width=0.85\linewidth]{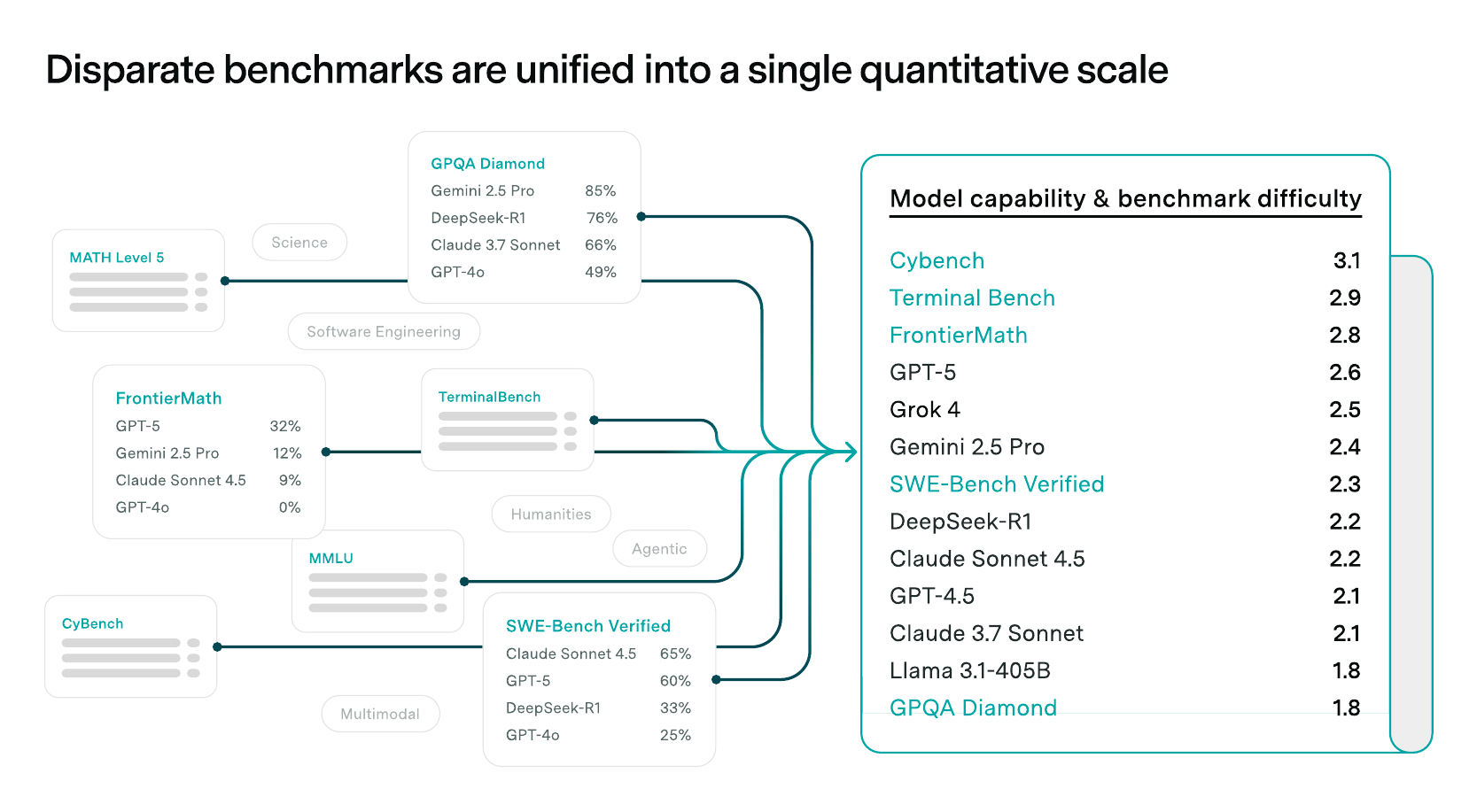}
    \label{fig:benchmark-stitching}
\end{figure}

Unlike some prior work on building aggregate capability metrics (\cite{kwa2025measuringaiabilitycomplete, ruan2024observationalscalinglawspredictability}), our framework involves minimal assumptions, e.g. it does not postulate any relationship between training compute or task length and model performance. This helps it flexibly capture what AI practitioners want to measure, and gives them the option to build new benchmarks to incorporate into the framework.
Relatedly, unlike approaches based on ELO scores (\cite{lmarena_leaderboard}) our framework does not require crowdsourcing data collection. This makes it extremely cheap to implement. 

Our framework opens several further possibilities. For one, it allows us to determine the relative difficulties of different benchmarks, not just model capabilities.\footnote{This is akin to how chess tactics puzzles have ELO scores, not just human players.} Moreover, by stitching together benchmarks with different difficulties, we can construct longer time-series of model capabilities -- these show a linear increase in estimated capabilities over time, despite our framework not incorporating any information about time or compute. 

We can then use these time-series to perform simple forecasts of future capabilities, and estimate the rate of algorithmic improvements, yielding results consistent with prior work. Finally, our framework makes it possible to detect rapid accelerations in model capabilities, a process which has been hypothesized to occur through the automation of AI R\&D (\cite{ai2027_research}). 

In this paper, we demonstrate all of these possibilities. The approach developed in this work is also applied in the \href{https://epoch.ai/benchmarks/eci}{Epoch Capabilities Index} with continuously updated data, using a rescaled normalization scheme for interpretability.

\section{Methodology}
\subsection{Model}
\label{subsec:model}
The core of our approach is to fit a statistical model that represents model capabilities and benchmark difficulties on a single numerical scale, similar in spirit to Item Response Theory (\cite{ColumbiaIRT}; see \Cref{appendix:relatedwork}). 

To estimate these capabilities and difficulties, we need to relate them to benchmark scores. We attempt to do this using the simplest statistical framework possible: we suppose that a model's benchmark accuracy scales roughly linearly with capability, unless the benchmark is too easy or too hard, in which case it gets 100\% or 0\% respectively.\footnote{Model performance could plateau below 100\% performance due to errors in the benchmark. It could also start above 0\% -- for example, four-option multiple choice question benchmarks typically have a baseline performance of 25\%, corresponding to uniform random guessing.} In practice, we expect the transitions from ``too easy" to ``linear scaling with capability" to ``too hard" to be relatively smooth, so we approximate this behavior using a sigmoidal relationship.\footnote{We relax this assumption in Appendix \ref{appendix:varying_statistical_model}.} Indeed, benchmark scores generally do show a roughly sigmoidal relationship with compute scaling and with time (e.g. see \cite{owen2024predictable}), so we choose a model that reflects this: 
\begin{equation}
    \text{score}(m, b) = \sigma(\alpha_b(C_m - D_b))
    \label{equation:main-model}
\end{equation}
Here \(\text{score}(m,b)\) represents the score that a model \(m\) would attain on a benchmark \(b\), which is bounded between 0\% and 100\%. The capabilities of the model are given by \(C_m\) and the difficulty of the benchmark is given by \(D_b\), and the difference between these two strongly influences benchmark scores. In particular, a difference \(C_m - D_b = 0\) corresponds to achieving a score of 50\%.\footnote{If the bounds are not 0\% and 100\%, then this may vary -- in general, this corresponds to the midpoint of the sigmoid.} Finally, the slope parameter \(\alpha_b\) controls the spread in difficulty of the tasks on benchmark \(b\).
 
This model rests on the assumption that AI model capabilities can be cleanly captured by a single number, which is of course not the case, but we opt for it nevertheless. 
The primary reason for this is simplicity: approaches to stitching together disparate benchmarks remain largely untested, and our objective was to observe whether the simplest formulation of this approach (with a single quantitative scale) would yield informative results. As we will see, even this simple approach yields results that appear consistent with expert assessments of model capabilities, as well as estimates of algorithmic progress in the literature (\cite{ho2024algorithmicprogresslanguagemodels}). This thus gives us confidence that our modeling approach is useful, and we welcome future work that builds upon this assumption (see Section \ref{sec:discussion}). 

\subsection{Data}
\label{sec:data}

To fit this model, we need a dataset of different \(\text{performance}(m,b)\) values, where our goal is to estimate \(\alpha_b\), \(C_m\), and \(D_b\) for all \(m\) and \(b\). We obtain this data from Epoch AI's collection of benchmarking data, which consists of data from two sources: 
\begin{itemize}
    \item \textbf{Internal data}: Benchmarking data from Epoch AI's internal evaluations, using consistent evaluation setups. These benchmarks generally include a fair degree of overlap, where many of the same models are evaluated across a large fraction of these internal benchmarks. Example internal benchmarks include FrontierMath (\cite{glazer2024frontiermathbenchmarkevaluatingadvanced}) and GPQA Diamond (\cite{rein2023gpqagraduatelevelgoogleproofqa}).
    \item \textbf{External data}: Benchmarking data that is stored in Epoch AI's benchmarking repositories, but collected from external sources. These generally do not use the same scaffold for different models on the same benchmark (though how much variation there is in the evaluation setup depends on the specific external benchmark). Example such benchmarks include MMLU (\cite{hendrycks2021measuringmassivemultitasklanguage}) and BIG-Bench Hard (\cite{suzgun2022challengingbigbenchtaskschainofthought}). 
\end{itemize}

In order to ensure that the model fits well, we further filter out models that have only been evaluated on \(\leq 3\) of the benchmarks. This means that the resulting data has more model overlap across different benchmarks, that help calibrate the estimates of \(C_m\) and \(D_b\).\footnote{We vary this assumption in Appendix \ref{appendix:robustness}, and find that our results remain very similar.}

In total, we fit the statistical model on 179 models and 38 benchmarks, with a total of 1324 performance scores. We provide more details about this data in Appendix \ref{appendix:data}.

\subsection{Optimization}
\label{sec:optimization}
We initialize the model capabilities \(C_m\) and benchmark difficulties \(D_b\) at 0, and the benchmark slopes \(\alpha_b\) at 1. We then fit Equation \ref{equation:main-model} using standard least squares regression using scipy's  \texttt{optimize.least\_squares} function, together with $L^2$ regularization with a default regularization strength of 0.1. This fitting process usually takes several seconds at most to complete, using the function's default optimization algorithm (Trust Region Reflective algorithm). 

During this stage we also address the model's identifiability issues, where multiple different sets of \(\alpha_b, C_m, D_b\) arrive at the same predicted \(\text{performance}(m,b)\). This occurs in two ways: 
\begin{enumerate}
    \item \textbf{Multiplicative rescale}: The model fits the data equally well with \(\{\alpha_b, C_m, D_b\}\) and a rescaled version \(\{k\alpha_b, \frac{C_m}{k}, \frac{D_b}{k}\}\) for some factor \(k \in \mathbb{R} \setminus \{0\}\).
    \item \textbf{Additive shift}: The model fits the data equally well independent of the absolute values of \(C_m\) and \(D_b\). So \(C_m + \delta\) and \(D_b + \delta\) would in theory yield just as good of a model fit, for some \(\delta \in \mathbb{R}\).
\end{enumerate}

These identifiability issues render it hard to compare the model predictions across different fits, e.g. with more or less data. To address this, we remove two degrees of freedom. We choose one benchmark (e.g. WinoGrande) and fix its slope \(\alpha_\text{WinoGrande} = 1\), and we then shift all estimates of \(C_m\) and \(D_b\) such that \(D_\text{WinoGrande} = 0\). Furthermore, we initialize all benchmark slopes to 1 and all estimated capabilities and benchmark difficulties to 0. 

\section{Results}
\label{sec:results}

\subsection{Determining model capabilities and benchmark difficulties}
\begin{figure}
    \centering
    \includegraphics[width=0.9\linewidth]{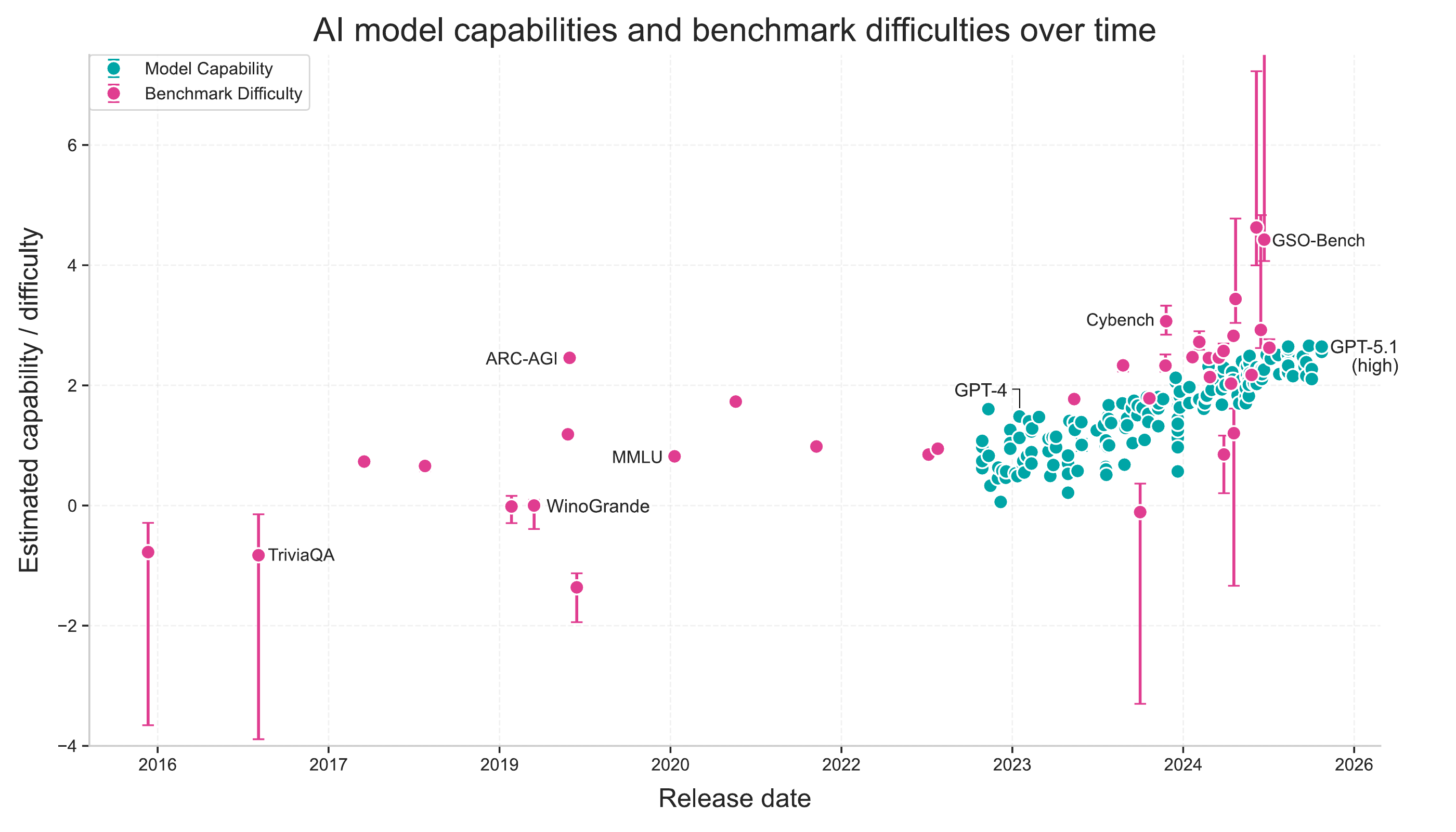}
    \caption{Estimated model capabilities \(C_m\) and benchmark difficulties \(D_b\) over time. 0 corresponds to the difficulty of the WinoGrande benchmark. We determine error bars through sensitivity analysis. Specifically, we perturb one model's capabilities and calculate the $L^2$ distance between the perturbed and original capability/difficulty sets. The error bars indicate the perturbation magnitude required to increase the loss by 5\%. We do the same with benchmark difficulties.}
    \label{fig:capabilities-and-difficulties-over-time}
\end{figure}

\begin{figure}
    \centering
    \begin{subfigure}[b]{0.49\textwidth}
        \centering
        \includegraphics[width=\linewidth]{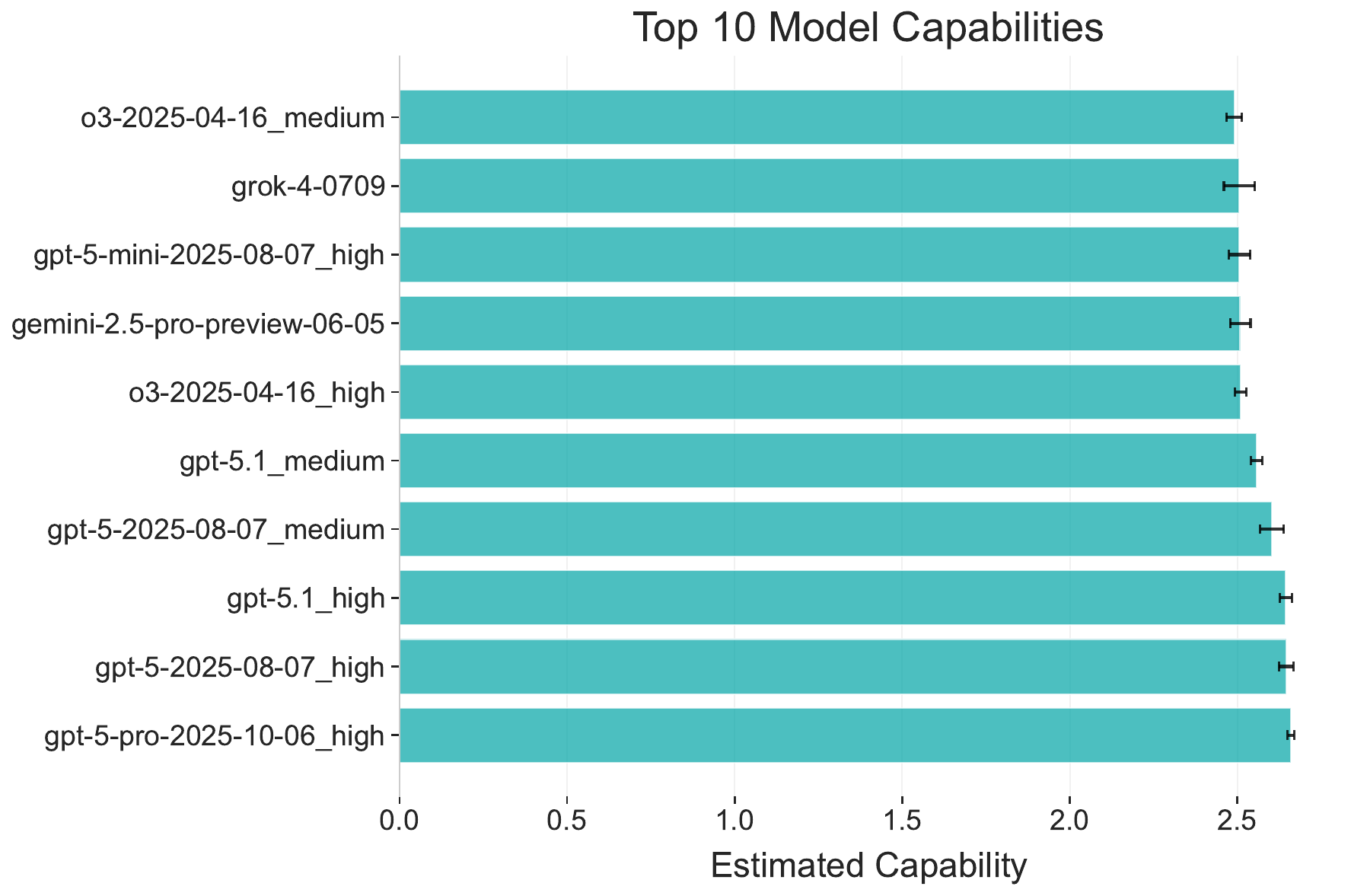}
        \caption{Model capabilities ranking}
        \label{fig:model-capabilities-ranking}
    \end{subfigure}
    \hfill
    \begin{subfigure}[b]{0.49\textwidth}
        \centering
        \includegraphics[width=\linewidth]{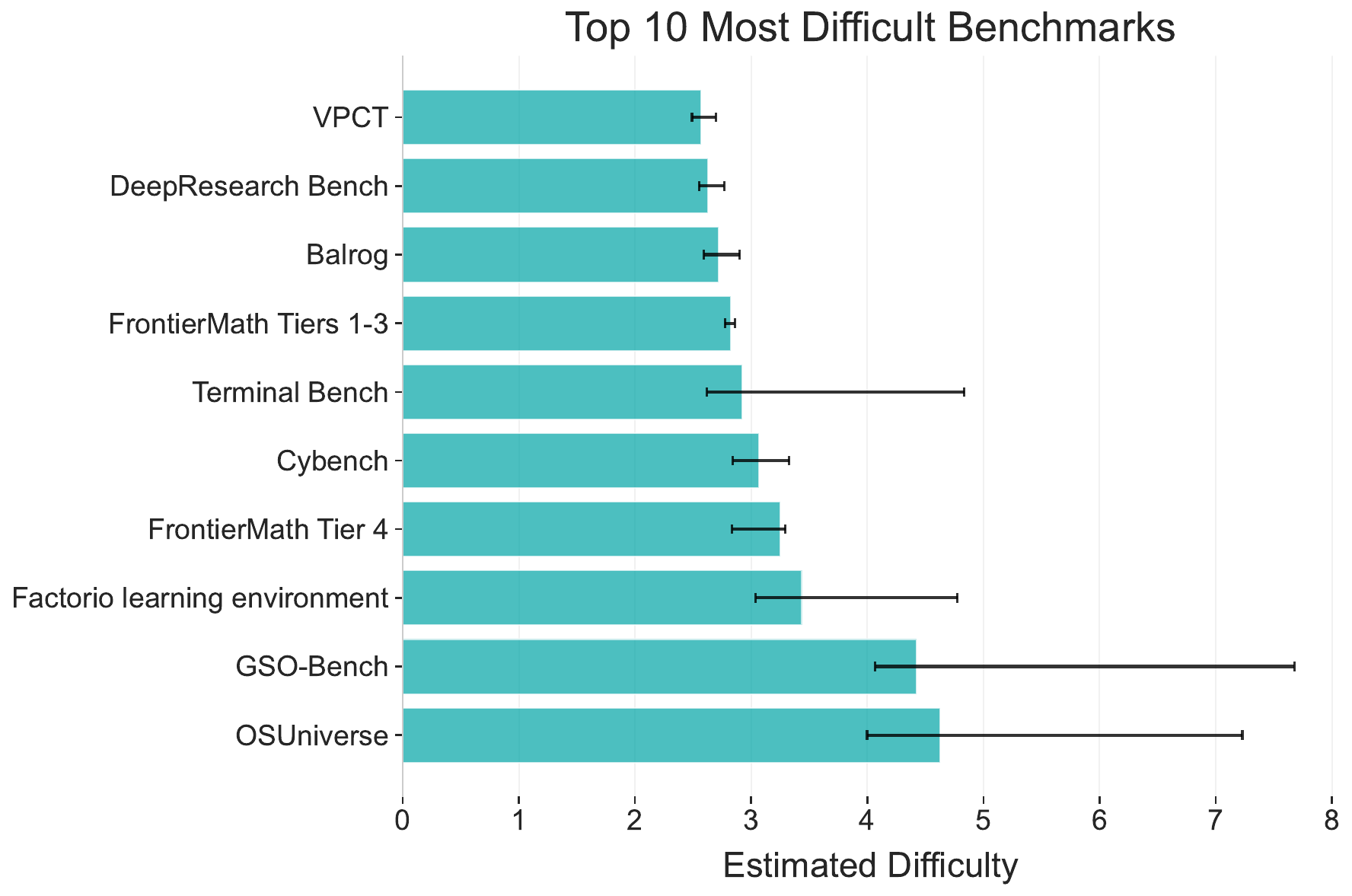}
        \caption{Benchmark difficulties ranking}
        \label{fig:benchmark-difficulties-ranking}
    \end{subfigure}
    \caption{When looking at the most capable models and most challenging benchmarks, our framework's predictions broadly match our intuitions from using models in practice. For example, the state-of-the-art model at the time of writing is GPT-5 or GPT-5.1.}
    \label{fig:combined-rankings}
\end{figure}

Fitting the model gives a set of model capabilities and benchmark difficulties -- we show how these vary with time in Figure \ref{fig:capabilities-and-difficulties-over-time}. This reveals that both the model capabilities and benchmark difficulties have been increasing over time. In order to validate that the model fit is reasonable, we also check the rankings more explicitly, and we show examples of this in Figure \ref{fig:combined-rankings}.

The first thing to note is that these rankings broadly match user experience of the relative capabilities of models. For example, Figure \ref{fig:model-capabilities-ranking} depicts recent models like GPT-5 as more capable than earlier models like o3, as one might expect.

Second, our approach also makes it possible to compare the relative difficulties of benchmarks. For example, Figure \ref{fig:benchmark-difficulties-ranking} illustrates how FrontierMath Tier 4 is substantially more challenging than VPCT. In fact, our estimates suggest that the most challenging benchmarks are still substantially harder than current state-of-the-art models are capable (see Figure \ref{fig:capabilities-and-difficulties-over-time}). That said, we think it is likely that these benchmark difficulties are somewhat overestimated. In particular, the newest benchmarks that we've included also tend to have few data points where models perform well. As a result, most of the data points tend to be at the ``flat part of the sigmoid" in Equation \ref{equation:main-model}, overestimating the benchmark's true difficulty.\footnote{This is very analogous to the issue pointed out in \cite{owen2024predictable}, where forecasting future language model performance is especially hard before the ``steep part of the sigmoid" is reached.} These benchmark difficulties also have much higher uncertainties, as indicated by the ``error bars" in Figure \ref{fig:benchmark-difficulties-ranking}.

\subsubsection{Interpreting estimated capabilities and difficulties}
\label{sec:interpretating_predictions}
So far we've primarily considered the relative model capabilities and benchmark difficulties, but how do we interpret this in a more absolute and quantitative way? Just looking at the estimated capability number isn't too informative, e.g. it's not clear what it means for GPT-5 to have an estimated capability of around 2.6. Instead, we think there are two main approaches for understanding the quantitative scale that follows from our framework. 

\textbf{Looking at capabilities differences relative to a pair of well-known models}. For example, many AI researchers have used both GPT-4 and GPT-5, and they can interpret capabilities in terms of ``GPT-4-to-GPT-5 jumps". Under this framing, o1-mini (high) is half a ``GPT-4-to-GPT-5 jump" better than GPT-4.

We look at models rather than benchmarks because these are probably more familiar to people, based on the models that they've used. However, there remain issues with this approach. For one, intuitions about AI models are often vague -- suppose future AI systems are two ``GPT-4-to-GPT-5 jumps" better than the current state-of-the-art. What tasks should we expect these AI systems to be able to perform? Moreover, it's not clear how much our intuitions of the difference in ``model capabilities" can be trusted. 

\textbf{Mapping estimated model capabilities to task time horizons}. Given the shortcomings of the prior approach, we also consider an alternative approach. This technique is based on \cite{kwa2025measuringaiabilitycomplete}, which looks at how long it typically takes humans to complete a task that AIs can complete at a certain success rate. For simplicity, we'll refer to this metric as the ``time horizon".

The idea is then to try and map the estimated model capabilities to the time horizons for different models - if the correlation seems sufficiently high, we can support the previous intuition with a more interpretable time horizon metric. We do this by looking at the models in \cite{kwa2025measuringaiabilitycomplete}'s time horizon data. We then randomly choose around 60\% of these models to fit a linear map from the estimated capabilities to the logarithm of the time horizon, using the remaining 40\% for validation. 

\begin{figure}
    \centering
    \includegraphics[width=\linewidth]{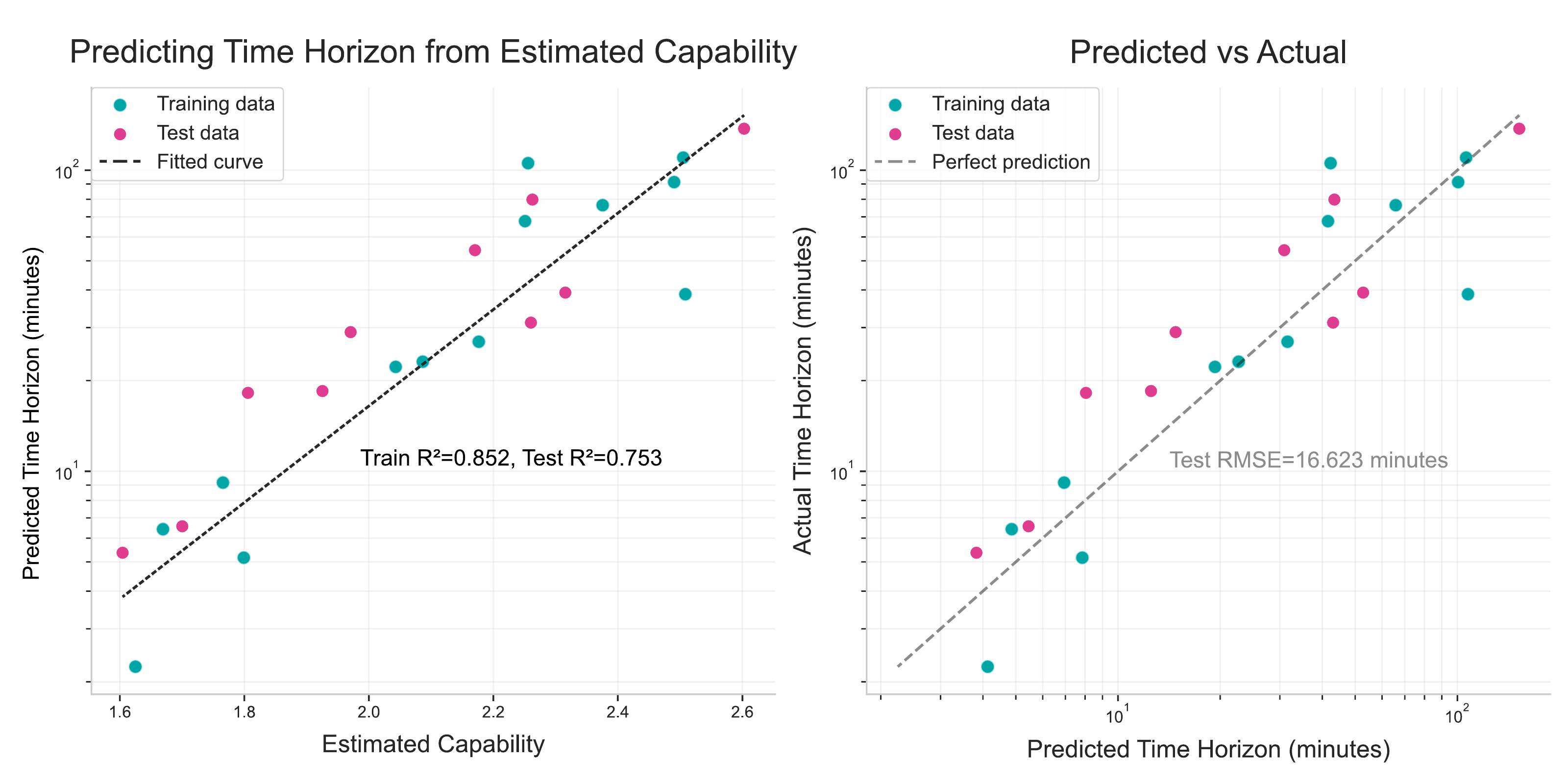}
    \caption{Our statistical framework and benchmark data predicts model time horizons from \cite{kwa2025measuringaiabilitycomplete} fairly well, with an \(R^2\) of 0.85. The resulting transformation is \(\text{time horizon} = \exp(3.69 \times C_m - 4.58)\).}
    \label{fig:predicting_time_horizon_from_capability}
\end{figure}

We find a high correlation between the estimated capability score and the log time horizon, with \(R^2 = 0.85\), as shown in Figure \ref{fig:predicting_time_horizon_from_capability}. Moreover, this approach of predicting time horizons outperforms an approach based on individual benchmarks 
For example, if we attempt to fit a linear model between the logit of GPQA diamond performance and the log time horizon, we instead get an \(R^2\) of 0.75.\footnote{This approach is similar to applying our framework but with a single benchmark, i.e. we set \(D_b = 0\) and \(\alpha_b = 1\), such that \(C_m\) is just \(\sigma^{-1}(\text{performance}(m,b))\).}
Doing this with FrontierMath yields a higher \(R^2\) of 0.97, but this might just be because there are substantially fewer models that overlap with the METR data (i.e. around 10 data points). With fewer data points, it becomes much easier to obtain very high correlations.
More generally, out of the 38 benchmarks that we consider in our analysis, 20 of the benchmarks do not have enough data to perform the analysis (i.e. there are fewer than 5 common datapoints between the METR time horizon data and the benchmark's data). Out of the remaining 18 benchmarks, the median \(R^2\) is 0.62.\footnote{6 of these models have a higher test \(R^2\) but they all have at most 13 data points, or around half of the number that is available from using the aggregate capability metric. Some benchmarks like SWE-Bench Verified and LiveBench are actively detrimental for predicting the time horizon, with a negative \(R^2\). }

Given that there seems to be a strong correlation, we can relate the estimated capabilities to the time horizon based on the transformation \(\text{time horizon} = \exp(3.69 \times C_m - 4.58)\). This gives us an alternative way of interpreting the trends we consider in Section \ref{sec:trends}. 

\begin{table}[h]
\centering
\begin{tabular}{cc}
\hline
\textbf{Model} & \textbf{Estimated capability (3 s.f.)} \\
\hline
GPT-5-high & 2.65 \\
o3-high & 2.51 \\
Grok 4 & 2.51 \\
o4-mini-high & 2.47 \\
o3-mini-high & 2.29 \\
o1-high & 2.28 \\
Claude Opus 4 (16K thinking) & 2.26 \\
Claude 3.7 Sonnet (64K thinking) & 2.22 \\
o1-mini-high & 2.16 \\
GPT-4.5 & 2.09 \\
Claude 3.5 Sonnet (Oct 2024) & 1.97 \\
Claude 3.5 Sonnet (June 2024) & 1.81 \\
GPT-4o & 1.77 \\
Claude 3 Opus & 1.67 \\
GPT-4o-mini & 1.66 \\
GPT-4 (Mar 2023) & 1.60 \\
Claude 2.0 & 1.40 \\
\hline
\end{tabular}
\caption{Example estimated model capabilities for notable models.}
\label{tab:model_capabilities}
\end{table}

\subsubsection{Are models optimized for different objectives?}
\label{sec:specialization}

Our framework assumes that model capabilities can be fully represented by a single number. However, different AI labs may be optimizing for different capabilities, such that some models may be better than others on some metrics, while being worse on others\footnote{More generally, different models may be optimized for different capabilities even within the same lab.}.

We can see this by inspecting our model's residuals: given just an estimated model capability, benchmark difficulty, and benchmark slope, what score does our framework predict the model will obtain on the benchmark, and how does this compare to the model's actual score? We plot predicted against actual scores in Figure \ref{fig:predicted_vs_actual}, and find that our framework makes good predictions on benchmarks with a sufficient number of data points: the average $R^2$ value is 0.71 for benchmarks with more than 25 scores, and 0.81 for benchmarks with more than 40 scores.

\begin{figure}
    \centering
    
    \includegraphics[width=\linewidth]{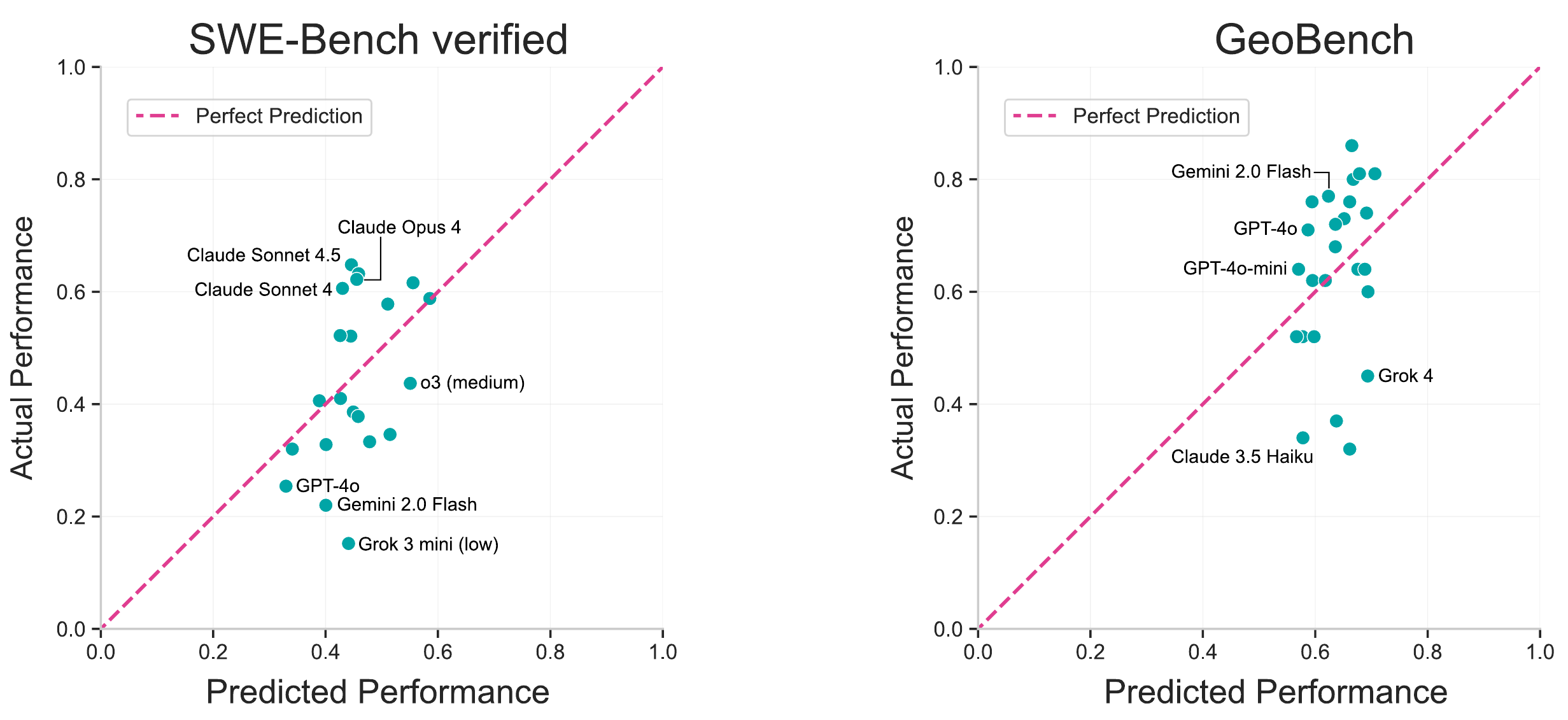}
    \caption{
    Predicting benchmark scores using our framework works well on most benchmarks, but there are some exceptions, two of which we show here. For example, Claude models perform better than predicted on SWE-Bench verified, a coding benchmark (left), but worse than predicted on a multimodal benchmark, GeoBench (right). This suggests some degree of specialization, where models from different labs are optimized for different objectives.
    }
    \label{fig:benchmark_comparison_swe_geo}
\end{figure}

Exceptions to this trend, i.e. benchmarks that our model fails to predict, can be especially interesting (see Figure \ref{fig:benchmark_comparison_swe_geo}). On the coding-heavy SWE-Bench verified benchmark, we see that the predicted performance is too high for models like Gemini 2.0 Flash, but too low for models like Claude Sonnet 4.5. In contrast, on the multimodal GeoBench benchmark we see Gemini models doing better than predicted and several Claude models doing worse than predicted. This provides weak evidence that Anthropic's models are more heavily optimized for code and less for multimodal tasks, whereas Google DeepMind's models are the reverse.

\subsection{Trends}
\label{sec:trends}

\subsubsection{Forecasting future capabilities}
A crucial benefit of our framework is that it makes it possible to compare models across a much wider range of capabilities. On the one hand, a benchmark may be far too challenging for a bunch of models to observe any difference in their performance (e.g. both GPT and GPT-2 would score 0\% on FrontierMath, despite the latter being substantially more capable). On the other hand, models may also be far too capable for the benchmark to detect any difference in their abilities (e.g. both o1-high and o3-high would likely ace the WinoGrande benchmark). Individual benchmarks may thus be unable to detect even major differences in model capability -- but being able to stitch together multiple benchmarks would ameliorate this problem. 

One particular case where this is especially applicable is in looking at longer-term trends in model capabilities. The fact that benchmarks only capture a limited range of difficulties manifests itself as rapid benchmark saturation (e.g. often within several months), due to the fast pace of model improvements. This poses issues for long-run extrapolation of model performance, but our statistical framework offers a potential way around this issue, helping us analyze trends and forecasts of future capabilities.

\begin{figure}
    \centering
    \includegraphics[width=\linewidth]{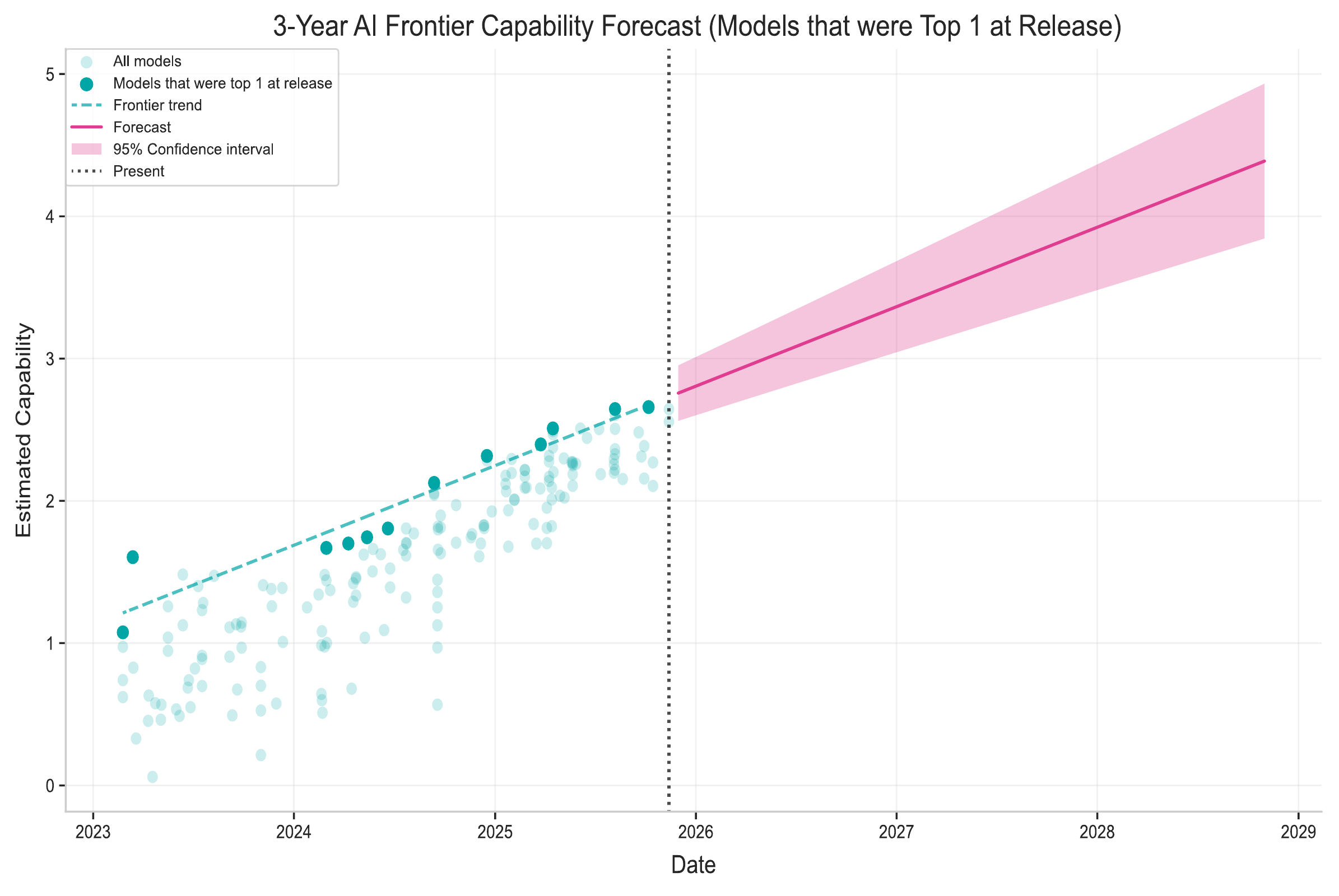}
    \caption{A naive forecast of future frontier model capabilities suggests a capabilities increase of 1.8 additional units over three years, double the improvement between GPT-4o and GPT-5 (high). Here, we forecast future frontier capabilities by extrapolating the capability scores of models that were state-of-the-art at release.}
    \label{fig:predicted-future-capabilities}
\end{figure}

\textbf{Frontier model capabilities have been improving at 0.55 capability units per year, similar to the capabilities gap between GPT-4.5 and GPT-5}. Consider the trend in model improvements in Figure \ref{fig:capabilities-and-difficulties-over-time}. If we look only at the capabilities frontier (i.e. models that push state-of-the-art capabilities at the time of release, see Figure \ref{fig:predicted-future-capabilities}), estimated model capabilities have been increasing at a rate of around 0.55 units per year, with a 95\% confidence interval of 0.45 to 0.67 units per year. This is similar to the size of the jump from GPT-4.5 to GPT-5-high, as shown in Table \ref{tab:model_comparison}, and roughly corresponds to a time horizon doubling time of around 5 months.\footnote{Following the approach outlined in Section \ref{sec:interpretating_predictions}, a capabilities increase of 0.55 units corresponds to time horizons increasing by \(\exp(3.69 \times 0.55 - 4.58) / \exp(3.69 \times 0 - 4.58) \approx 7.6\) times per year. This corresponds to a doubling time of \(\log(2)/\log(7.6) \times 12 \approx 4\) months.}

One important implication of this is that several frontier labs are quite close at the frontier of capabilities. In particular, we know from Figure \ref{fig:model-capabilities-ranking} that the top 10 models are within around 0.2 capability units of each other, including models from OpenAI, Google DeepMind, and xAI. So if we believe this 0.55/year capability trend, this would suggest that these labs are all easily ``within a year of each other", and more likely on the order of a few months.

\textbf{Extrapolating the trend in capabilities growth suggests a capabilities score of around 4.4 units by October 2028}. With a trend of 0.55 capability units per year, a naive extrapolation suggests an increase of around 1.65 units by October 2028, or three years from the time of writing. This is equivalent to three GPT-4.5 to GPT-5-high jumps in capabilities over the existing state-of-the-art, or two GPT-4o to GPT-5 jumps (see Table \ref{tab:model_capabilities}). Alternatively, we can interpret this in terms of the map to the time horizon metric, corresponding to a time horizon of 17 weeks (95\% CI: 2.6 to 30 weeks).\footnote{One could also use this extrapolation to perform forecasts on individual benchmarks, especially on those that are have fewer existing model evaluations. Though in this case it is especially important to check that the issue of capabilities being multidimensional (see Section \ref{sec:specialization}) isn't a major issue on these benchmarks. In such cases it may be more valuable to just directly extrapolate the data on the benchmark alone.} 

Of course, this naive forecast may turn out to be incorrect. For one, this approach to forecasting doesn't account for the drivers of the trend, and it's possible that things might slow it down. For example, there might be a slowdown in compute scaling (\cite{epoch2025computescalingwillslowdownduetoincreasingleadtimes}) that reduces the growth of training compute, and hence model capabilities. Alternatively, AI systems could contribute to the process of AI R\&D, potentially accelerating algorithmic progress and thus model capabilities growth (\cite{eth2025willairdautomati, erdil2024estimating}).

\begin{figure}
    \centering
    \includegraphics[width=\linewidth]{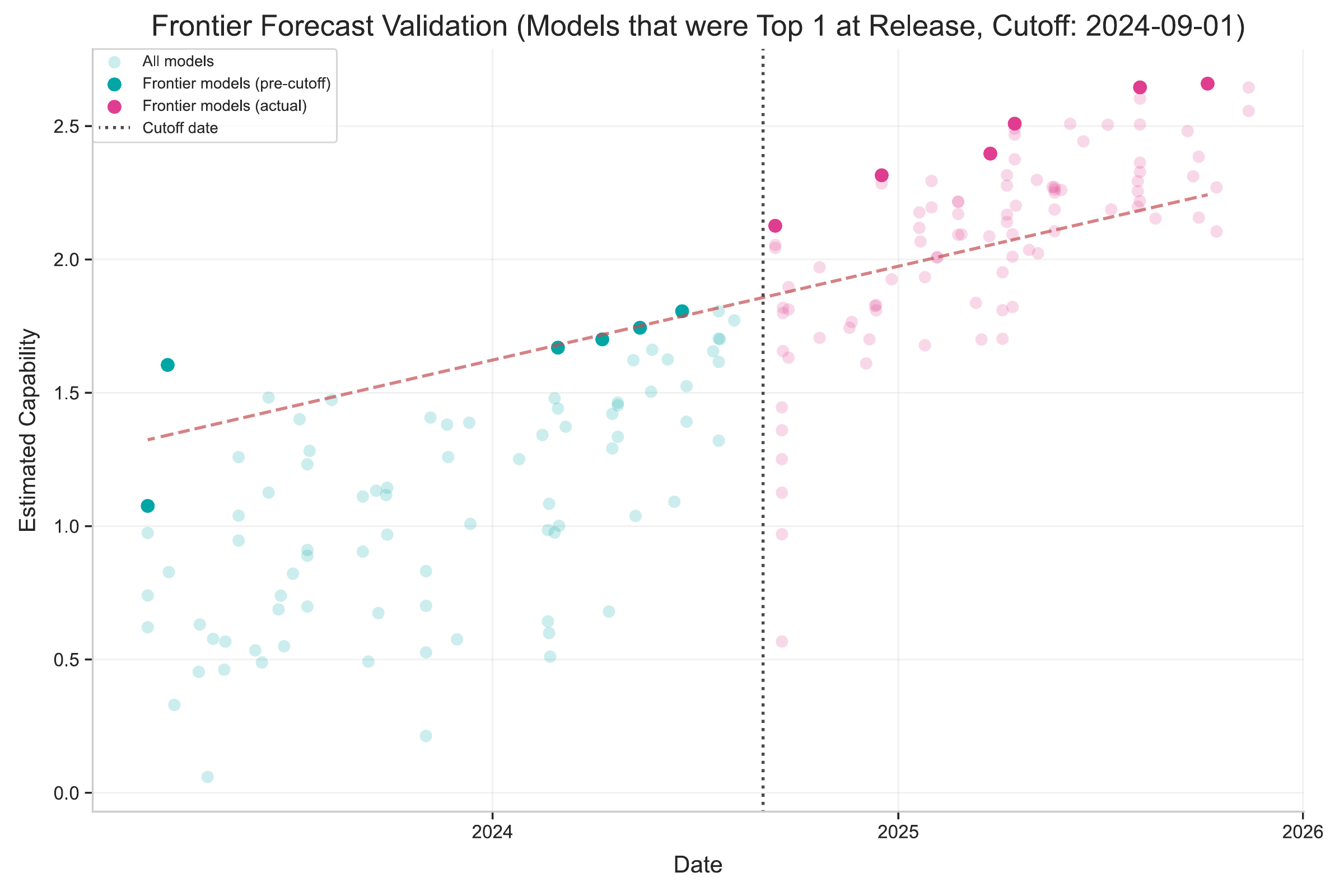}
    \caption{Linearly extrapolating the trend in frontier model capabilities at the start of September 2024 would have slightly underestimated model capabilities over the next year, likely due to rapid and widespread adoption of reasoning models at the time.}
    \label{fig:frontier_forecast_validation}
\end{figure}

There may have already been changes in the trend in model capabilities over time. 
For example, if we had attempted to perform a forecast of forthcoming model capabilities in September 2024, we would have slightly underestimated the rate of progress, as shown in Figure \ref{fig:frontier_forecast_validation}. This is likely because of the rise of reasoning models, which represented a notable algorithmic improvement and paradigm shift (\cite{epoch2025quantifyingthealgorithmicimprovementfromreasoningmodels}). Notably, all of the models that pushed frontier estimated capabilities after the October 2024 cutoff are reasoning models (e.g. o1, o3, and Gemini-2.5 Pro). 
Choosing this start date for our forecast  is in some sense maximally pessimistic -- it was deliberately chosen to occur right at the introduction of reasoning models. But in general it's certainly possible that such accelerations occur in the future if there's a change in the drivers of the trend, which could make our naive forecast inaccurate. 

\subsubsection{Algorithmic progress}
\label{sec:algorithmic-progress}

Another use case of our approach is to develop long time-series of model performance, which we can use to analyze the rate of algorithmic progress. 
In particular, prior work on this subject (\cite{ho2024algorithmicprogresslanguagemodels, erdil2023algorithmicprogresscomputervision}) have generally been constrained by the availability of long-run benchmark data (since most recent benchmarks tend to saturate within one to two years), and this approach allows us to bypass this constraint to some extent. 

\begin{figure}
    \centering
    \includegraphics[width=\linewidth]{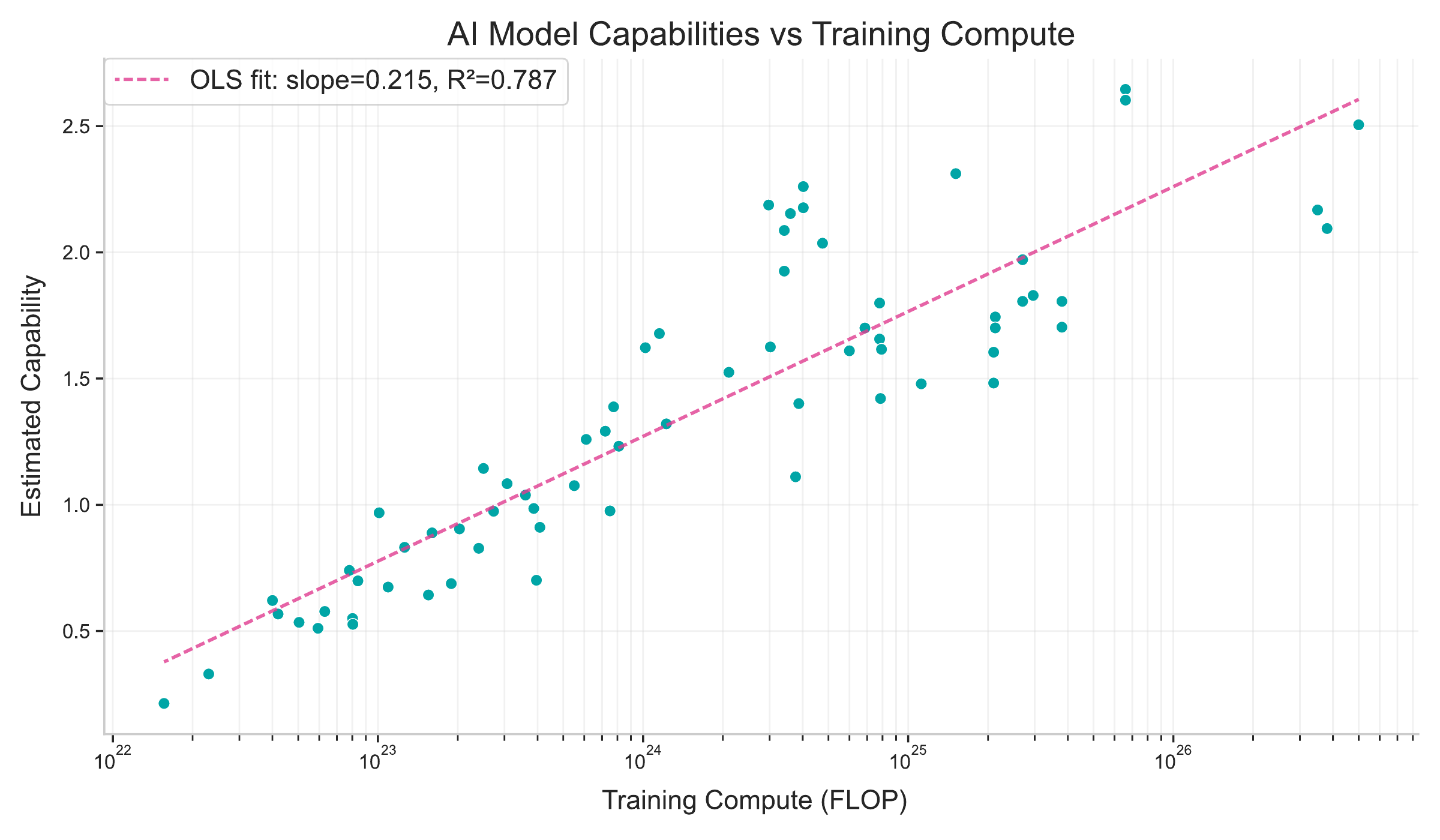}
    \caption{Estimated model capabilities appear to grow linearly with log training compute.}
    \label{fig:capability_vs_log_compute}
\end{figure}

To estimate the rate of algorithmic progress, we combine the estimated model capabilities together with data on training compute and model release dates. We show this data in Figure \ref{fig:capability_vs_log_compute}.

As in prior work, we define algorithmic improvements in terms of decreases in the training compute needed to attain a certain level of performance, or in this case estimated model capability (\cite{ho2024algorithmicprogresslanguagemodels, erdil2023algorithmicprogresscomputervision}). 

We then estimate algorithmic progress using a statistical model relating estimated model capabilities \(C_m\) to the training FLOP \(F_r\): 
\begin{equation}
    C_m = k \log F_m + b \iff F_m = \exp\left(\frac{C_m}{k}\right) \exp\left(-\frac{b}{k} \right).
    \label{eq:algorithmic_progress}
\end{equation}
Here \(k\) is a parameter that determines how model capabilities \(C_m\) scale with increases in training compute. \(b\) helps capture algorithmic improvements -- the higher the value, the better the ``algorithmic quality" of a model, since it means less compute \(F_m\) is needed to achieve the same \(C_m\). Our goal with this approach is to measure algorithmic progress of the form where \(b\) increases over time at the same \(k\). This involves two steps: 
\begin{enumerate}
    \item \textbf{Estimate \(k\) for specific model families}: We first identify several model families which use the same training recipe (algorithms and data).\footnote{We restrict models to using the same training recipe since this allows us to isolate out the impact of algorithmic progress, estimating \(k\) on a set of models with the same algorithmic quality. This helps us overcome an issue that is quite common in work attempting to estimate algorithmic progress, which is that the high correlation between training compute and algorithmic innovations makes it hard to attribute capabilities improvements to each (\cite{ho2024algorithmicprogresslanguagemodels}).}
    Within each family, we then plot \(C_m\) vs \(\log F_m\), which we expect to be linear, and allowing us to estimate \(k\) from the slope of the fitted line. We then take the average of the estimated \(k\) values across the families to obtain a single \(k\).
    \item \textbf{Given this \(k\), estimate \(b\) for all models and back out a rate of algorithmic progress}: With estimates of \(b\) for each model, we can then work out the rate of change \(\frac{db}{dt}\) among models that were at the frontier of capabilities. Over one year, the change is then \(\Delta b\), and the decrease in training FLOP needed is then \(\exp\left(\frac{\Delta b}{k}\right)\).
\end{enumerate}

In practice we have sufficient data on three families of models: LLaMA, LLaMA 2, and LLaMA 3.1 -- this yields an average \(k = 0.168\) across the families.\footnote{We determine this average by taking a weighted average of the estimates of \(k\) in each family, where the weighting is done by the number of models in each family. 
In our case, each family has four models, so the weighted average is the same as taking an arithmetic mean.} Among non-distilled models at the frontier of model capabilities,\footnote{We drop distilled models from the dataset since we are interested in capturing the relationship between model capabilities and training compute for the final training run. This relationship might be heavily influenced by additional compute sources, such as from distillation or substantial quantities of synthetic data generation (\cite{epoch2025threeissuesunderminingcomputebasedaipolicies}).} we see \(\frac{\Delta b}{\text{1 year}} = 0.297 \). So as a central estimate, the training compute needed to reach a certain capability has been declining at \(\exp(\frac{0.297}{0.168}) \approx 6\times\) per year. This also implies that even if we held training compute fixed, algorithmic progress alone could still result in a capabilities increase of around 0.3 units per year. 

However, these estimates are highly uncertain, in large part due to the uncertainty in \(k\). Across the three families, we end up with a very wide range of numbers, from 1 to 50\(\times\) reduction in training compute requirements per year. Similarly, the fixed-compute increase in capabilities is around 0.1 to 0.5 capability units per year.

We also consider the rate of algorithmic progress when we look at the ``frontier of algorithmic quality". We can think of this in terms of the graph of \(b\) against time --- since \(b\) captures algorithmic quality of the models, with larger values corresponding to better algorithms, we consider the ``frontier in algorithmic quality" as the models that have the highest \(b\) values at any particular point in time. If we then look at \(\frac{db}{dt}\) for these frontier models, we find a substantially higher central estimate of the rate of algorithmic progress, at around 9\(\times\) per year, with estimates ranging from 3 to 40\(\times\). We show the full results in \ref{tab:llama_progress}.

Much of this uncertainty comes from the uncertainty in the estimate of \(k\). In particular, we get a much lower estimate of \(k\) with LLaMA 3.1 models compared to LLaMA and LLaMA 2 models. This leads to a much higher estimated rate of algorithmic progress,\footnotemark substantially higher than central estimates in prior work (\cite{ho2024algorithmicprogresslanguagemodels, whitfill2025forecastingaitimehorizon}). 
\footnotetext{Intuitively, a lower value of \(k\) means that compute is less helpful for increasing capabilities, so to explain observed capabilities improvements, the model puts more weight on rapid compute efficiency improvements.}
However, it's not clear to us why this is the case -- to our knowledge, smaller LLaMA 3.1 models were not distilled. So we are more suspicious of the estimates obtained from LLaMA 3.1, but include it here since we do not have a strong reason to exclude it. 

\begin{table}[h]
\centering
\begin{tabular}{lcccccc}
\hline
\textbf{Model Family} & \(k\) & \(\Delta b\) [95\% \textbf{CI}] & \multicolumn{2}{c}{\textbf{Annual compute efficiency gain} [95\% \textbf{CI}]} \\
\cline{4-5}
& & & \textbf{Model capabilities} & \textbf{Algorithmic quality} \\
\hline
LLaMA & 0.18 & 0.30 [0.12, 0.48] & 5\(\times\) [2, 10] & 8\(\times\) [6, 10] \\
LLaMA 2 & 0.20 & 0.26 [0.05, 0.47] & 4\(\times\) [1, 10] & 4\(\times\) [3, 6] \\
LLaMA 3.1 & 0.12 & 0.40 [0.24, 0.49] & 20\(\times\) [7, 50] & 30\(\times\) [20, 40] \\
\hline
\end{tabular}
\caption{Parameter estimates and rates of algorithmic progress across LLaMA model families, comparing the frontier in model capabilities versus algorithmic quality. We derive 95\% confidence intervals for these by looking at the slope of \(b\) vs \(t\), and looking at the associated t-statistic and standard error.}
\label{tab:llama_progress}
\end{table} 

Besides having wide uncertainty intervals, this approach various other shortcomings. Most crucially, algorithmic progress could also involve changing \(k\), such that the rate of algorithmic progress might depend on the specific scale of compute under question -- for example, the rate of reduction in compute requirements might be faster at \(10^{25}\) FLOP compared to \(10^{21}\) FLOP. Instead, the model in Equation \ref{eq:algorithmic_progress} assumes that these two rates are equal. Unfortunately, we do not have sufficient data to test this scale-dependence of algorithmic progress in detail, so for the purposes of this paper we present our results assuming scale-independence. We discuss additional algorithmic progress experiments in Appendix \ref{appendix:algorithmic-progress}.

\subsection{Detecting rapid performance accelerations}
\label{sec:detecting-acceleration}

One scenario that has been repeatedly raised as a possibility for future AI progress is that of rapid performance accelerations (see for instance \cite{ai2027_research}). Looking only at ordinary benchmarks may fail to capture this because we do not know the benchmark slope \(\alpha_b\), and we also do not know how difficult different benchmarks are relative to one another -- simply observing substantial benchmark saturation in quick succession may thus be somewhat misleading.\footnote{It may also be unclear which benchmarks to pay attention to, but our framework also has this same weakness, since the estimated capabilities depend on the choice of benchmarks.} 

\subsubsection{Synthetic data experiments}

In this section, we run synthetic data experiments to test whether our model can detect rapid capability accelerations. As a concrete example, we ask how much data is needed to detect a 2\(\times\) acceleration in capabilities. 

To do this, we use synthetic data to simulate a scenario in which model capabilities suddenly accelerate, while benchmark difficulties improve linearly. We first generate synthetic data on two trends. One trend corresponds to improvements in model capabilities, which we model as a piecewise linear trend with an acceleration in a given year. The second trend corresponds to improvements in benchmark difficulties, which is simply linear over time.\footnote{We illustrate these two trends in Figure \ref{fig:capabilities-difficulties-trend-broad}.} These are used to determine a distribution of benchmark scores across different benchmarks over time, which constitutes the data that can be used to detect the capabilities acceleration (the underlying capability scores and benchmark difficulties are not observable).

To test whether or not the framework can detect an acceleration of model capabilities, we fit a piecewise linear model with a single breakpoint, focusing primarily on frontier models. We then compare the slopes of the model before and after the breakpoint, and if the ratio of the slopes (i.e. \(\frac{\text{post-cutoff slope}}{\text{pre-cutoff slope}}\)) exceeds a certain threshold, we consider an acceleration as ``detected". Note that the breakpoint is also estimated rather than manually specified.

\begin{figure}[t]
    \centering
    \includegraphics[width=\linewidth]{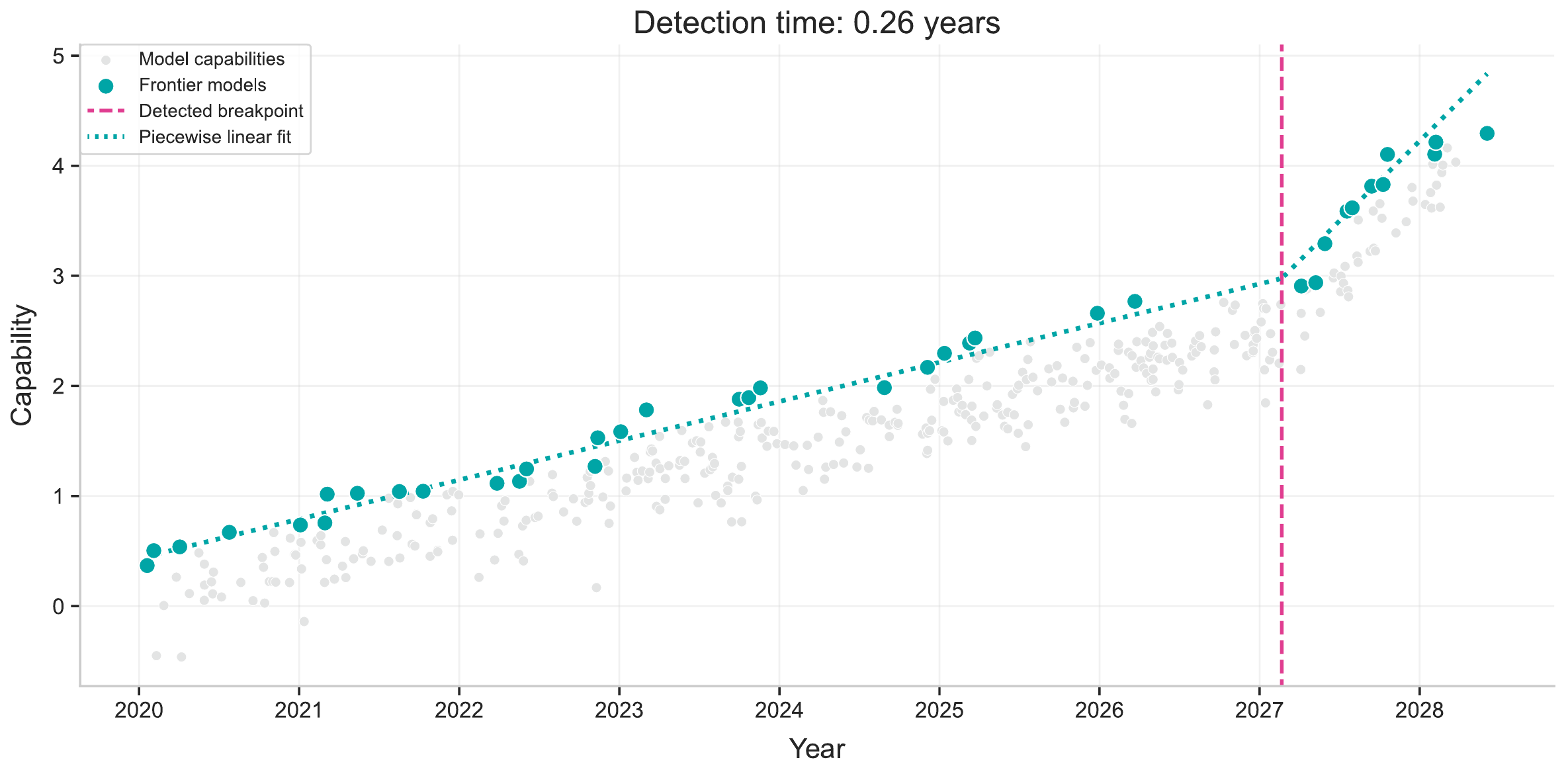}
        \caption{Our detection pipeline focuses on frontier models, and identifies a 2\(\times\) acceleration around 3 months after the actual point of acceleration in 2027.}
        \label{fig:synthetic-data-narrow-detect}
\end{figure}

One example of this is illustrated in Figure \ref{fig:synthetic-data-narrow-detect}. This considers a world where models exhibit a 2\(\times\) acceleration in capabilities growth. Focusing on frontier models, our framework has enough data to detect such an acceleration around three months after it starts. 

We test how sensitive our detection method is by sweeping over acceleration magnitude (\(2\times, 4\times, 8\times\)), noise levels (\(0.5\times\) to \(4\times\) a baseline level), and the fraction of models exhibiting acceleration (25\% vs 100\%). 

Our framework can reliably detect accelerations within 2--3 months under realistic conditions. Under moderate noise (\(1\times\) baseline), a \(2\times\) acceleration is detected within 2--3 months when all models accelerate, while higher accelerations (\(4\times\) and \(8\times\)) are detected slightly faster. However, noise level dominates detection performance: increasing noise from \(0.5\times\) to \(4\times\) increases the detection time by \(4.1\times\) on average. The fraction of models exhibiting acceleration has the second-largest effect, with universal acceleration (100\% of models) enabling 27\% faster detection than sparse acceleration (25\% of models). Surprisingly, acceleration magnitude itself only has a small impact in our simulations---a \(4\times\) acceleration is detected only slightly faster than \(2\times\) or \(8\times\).\footnote{More details about our synthetic data generation are described in Appendix~\ref{appendix:synthetic-data}.}

To validate detection specificity, we test false positive rates by generating synthetic data with no acceleration. The false positive rate averages 38\%, ranging from 14\% to 54\% depending on noise levels and observation windows. These relatively high rates indicate that our detection approach is useful as a monitoring tool to flag potential accelerations for further investigation. However, it should be combined with other evidence rather than treated as definitive on its own.\footnote{Detailed false positive analysis is provided in Appendix~\ref{appendix:synthetic-data}.}

\subsubsection{Detecting acceleration in the historical data}

\begin{figure}
    \centering
    \includegraphics[width=\linewidth]{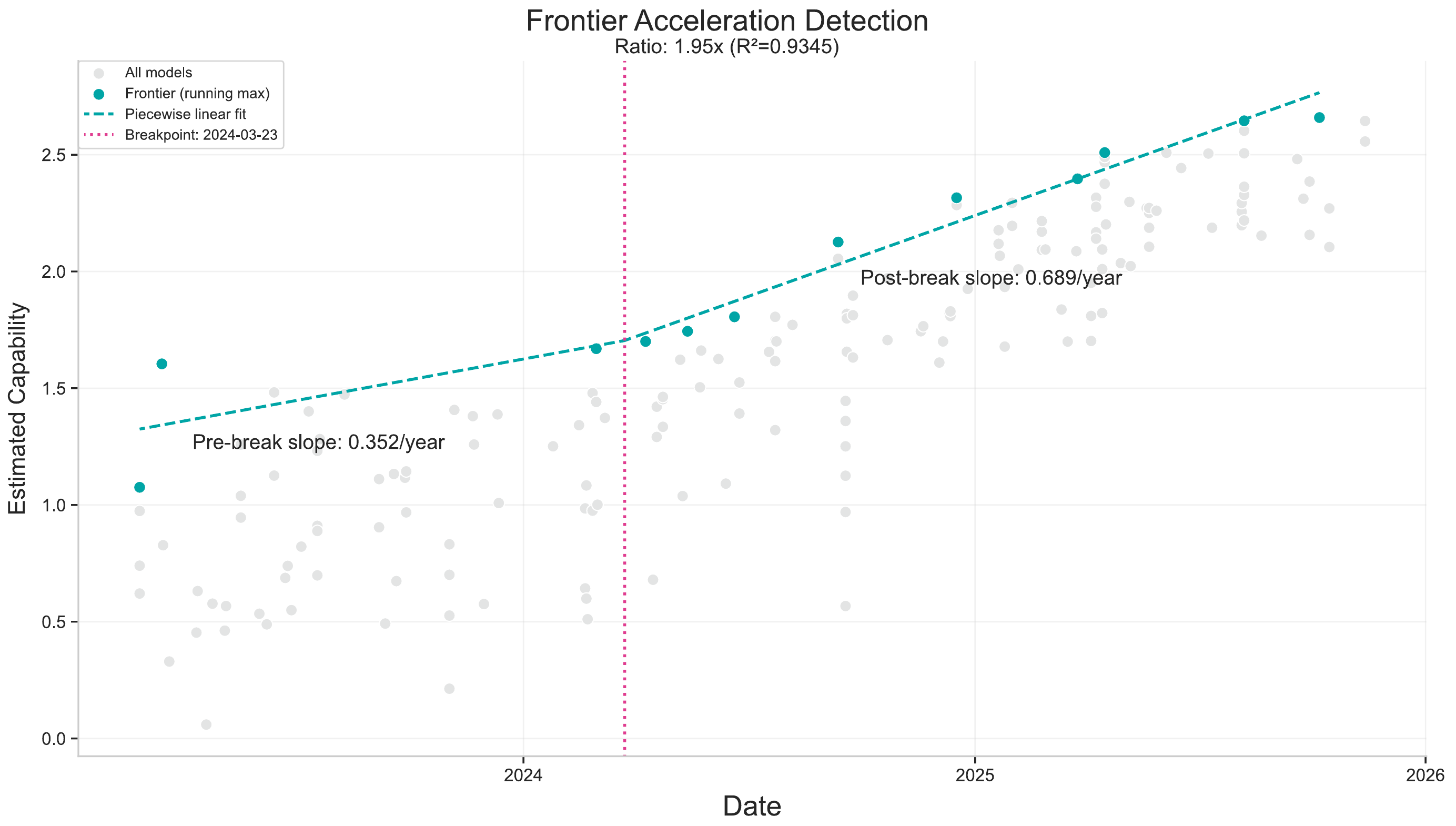}
    \caption{Running our detection pipeline on real data finds a noticeable \(1.95\times\) acceleration in frontier model capabilities, with a breakpoint several months before the rise of reasoning models (March 2024).}
    \label{fig:real_data_acceleration_detection}
\end{figure}

The above results on synthetic data naturally raise the question of whether our framework would have detected such an acceleration in the historical data. Running this pipeline, we observe a breakpoint in April 2024, with a pre-break slope of 0.352/year and a post-break slope of 0.689/year. The acceleration is thus 1.95\(\times\), which is close to but does not cross the 2\(\times\) threshold that we consider here. We show the results of this fit in Figure \ref{fig:real_data_acceleration_detection}.

This breakpoint is consistent with findings from \cite{kwa2025measuringaiabilitycomplete} on time horizons (see Section \ref{sec:interpretating_predictions}). In particular, \cite{kwa2025measuringaiabilitycomplete} identify a potential acceleration in time horizon doubling times from 7 months to 4 months in 2024. This corresponds to an acceleration of around 7/4 = 1.75\(\times\), which is similar to the acceleration estimated using our approach.

\section{Discussion}
\label{sec:discussion}

Our work has several limitations that point towards potential directions of further research. 

\subsection{Challenges in interpreting model capability and benchmark difficulty scores}

It is unclear what exactly is meant by a ``capabilities score" of 3. Even if we repeatedly draw analogies to prior model improvements (e.g. the size of the jump between GPT-4o and o1-high), our intuitions for model capabilities may be based on factors beyond raw benchmark performance, such as speed and price, and these are completely omitted from our model. There are other approaches for interpreting these scores, for example by relating them to time horizons as outlined in Section \ref{sec:interpretating_predictions}, but these are also fairly speculative. Fortunately, many applications of our approach do not require an intuitive interpretation of the estimated capability scores -- e.g. when estimating algorithmic progress (see Section \ref{sec:algorithmic-progress}).

\subsection{Limitations inherited from individual benchmarks}

Our framework builds upon existing benchmark results. This means that it inherits many of the limitations of modern AI benchmarks. For example, benchmarks often lack realism and are far from covering the full spectrum of economically useful tasks. The set of benchmarks that we used, while diverse, is likely biased toward specific domains such as mathematics and reasoning tasks. Some benchmarks also have a high error rates (\cite{epoch2025whatskillsdoesswebenchverifiedevaluate, epoch2025whatdoesosworldtellusaboutaisabilitytousecomputers}), which imposes a ceiling on the maximum achievable benchmark score. Finally, benchmark scores often depend heavily on the specifics of the evaluation setting (for example, the prompt, scaffold, or amount of inference compute), which are often different from the setups used when deploying AI products in the real world. These issues mean that the estimated model capability scores ($C_m$) may not fully represent general, real-world capabilities. 

Fortunately, our methodology is flexible: practitioners can mitigate these issues by applying it to a curated set of benchmarks, or by assigning greater weight to benchmarks that they find more reliable or relevant to their preferred applications. In practice, the main constraint remains the need for sufficient data: benchmarks must have enough models evaluated on them, with enough overlap of models across different benchmarks for this ``stitching'' framework to function.

\subsection{AI capabilities cannot be fully described by a single number}

As described in \cref{subsec:model}, our model explicitly assumes that model capabilities can be described by a single number $C_m$. In practice, model capabilities likely depend on a range of different skills, so that while Model A might be superior to Model B in one domain (e.g., coding), the ranking could be reversed in another domain (e.g., multimodal understanding). The results in \cref{sec:specialization} show such examples of model specialization, but the overall trend is that the model capability scores $C_m$ are good predictors of benchmark scores. An interesting direction for future work would be to build multidimensional analogues of our framework, and to investigate whether they reveal distinct latent components of model ability.

One example of this is \cite{epoch2025benchmarkscoresgeneralcapabilityclaudiness}, which performs a principal components analysis using essentially the same data source as we do in this paper. This finds that two principal components explain most of the variance in benchmark scores, with the first corresponding roughly to what our framework captures, and the second corresponding to ``Claudiness" (i.e. strong coding abilities, but poor vision and math abilities --- a skill profile common in Claude models).

\subsection{Our model uses average benchmark scores rather than results on individual questions}

Unlike Item Response Theory (IRT), which operates on pass/fail data for individual questions (\cite{ColumbiaIRT}), in this work we take aggregate benchmark scores as our primary input. For instance, for a benchmark like MMLU, we use the model's average accuracy across all $\approx 16,000$ questions, as opposed to the result for each specific question. We made this choice because item-level data is not available for many benchmarks where we could nevertheless average scores, and we wanted to prioritize incorporating a wide diversity of tasks and difficulties.

However, this decision means that our approach does not verify some desirable invariance properties. For example, if a benchmark $B$ is arbitrarily split into two benchmarks $B_1$ and $B_2$ with similar difficulties, fitting the model with $B_1$ and $B_2$ is essentially tantamount to giving $B$ double the weight, despite this splitting being an arbitrary transformation that does not change the underlying question-level data. Therefore, we believe that a promising future direction is obtaining item-level results across a wide variety of high-quality benchmarks, and adapting our approach at the question level.

\subsection{Future directions for detecting capability accelerations}
The main output of our methodology is a long time-series of unified model capability scores. In \cref{sec:results}, we've presented some applications of this series, including detecting capability accelerations. However, the detection method we employed remains fairly \textit{ad hoc} and resulted in a high false positive rate on synthetic data. More principled methods drawing from the time series or sequential testing literatures might improve the specificity and reliability of acceleration detection.

\section{Conclusion}
In conclusion, in this paper we presented an approach that in effect allows us to ``stitch together" different benchmarks, compressing both model capabilities and benchmark difficulties down into a single numerical scale. 

The resulting data allows us to perform naive forecasts of future capabilities in a way that was not possible with individual benchmarks, and allows us to estimate a rate of algorithmic progress that is broadly consistent with prior work. We also demonstrate how future data could be used to spot accelerations in model capabilities over time, which might otherwise be hard to detect. 

\textit{Acknowledgements: We thank Luke Emberson, Brendan Halstead, Anna Wang, David Owen, Eli Lifland, Ryan Greenblatt, Fabien Roger, Alexander Barry, Tom Cunningham, Luca Righetti, Parker Whitfill, Evgenii Opryshko, Matt Kowal and Nate Rush for their feedback and support.}

\newpage
\printbibliography
\newpage
\appendix
\section*{Appendix}
\section{Related work}
\label{appendix:relatedwork}
\textbf{Observational scaling laws}. \cite{ruan2024observationalscalinglawspredictability} propose an approach to estimate language model performance, without having to perform expensive scaling experiments. They propose that language model capabilities can be decomposed into a low-dimensional space with a handful of principal components, which explain much of the observed variance in benchmark performance, and scale log-linearly with training compute within certain families of models.  

\textbf{Time-horizon based capability metrics}. A core motivation for our framework is that it allows us to obtain long-run time series of model capabilities. 
However, we are not the first to attempt this. 
Most notably, \cite{kwa2025measuringaiabilitycomplete} propose a new metric for AI capabilities: the time humans typically take to complete tasks that AI models can complete with a certain success rate. 
The idea is that as AI systems improve, they become able to do tasks that take humans longer to do. 
This metric was estimated on a set of software-related tasks (\cite{rein2025hcasthumancalibratedautonomysoftware}). 
In contrast, our framework is more flexible and cheaper to implement, but it is also harder to directly interpret. 

\textbf{ELO rankings for models}. Besides benchmark scores, there have also been attempts at compressing model capabilities into a single scale, such as through ELO score-based leaderboards. Notable examples of this include the LMArena (\cite{lmarena_leaderboard}) and SciArena (\cite{sciarena_2025}). 
But there are two crucial differences compared to our approach. First, while ELO-based approaches generally arrive at rankings through repeated human votes, our approach does not require any additional evaluation effort. Instead, it just aggregates the results of already-performed benchmark evaluations, thus it is substantially less costly to implement. Second, our work allows us to compare the difficulties of benchmarks, not just model capabilities. 

\textbf{Predicting benchmark performance}. Various works have investigated the predictability of downstream benchmark performance. 
For instance, \cite{owen2024predictable} examines the relationship between training compute and language model benchmark performance, finding that performance is moderately predictable from compute scaling. 
\cite{schaeffer2025predictingdownstreamcapabilitiesfrontier} further point out that this predictability depends on the specific metric used to measure benchmark task performance. 

\textbf{Item response theory}. Item response theory (IRT, see \cite{ColumbiaIRT} for an example reference) is a statistical framework that attempts to explain how well people answer test questions, based on an underlying ``ability" of test-takers, and on ``difficulty" of the test questions (``items"), among other factors.
The underlying statistical model is very similar to the one we outline here, with the important difference that IRT looks at item-level data, whereas we consider overall benchmark scores. 
Indeed, since we can use benchmark scores even when item-level data is not available, we are able to aggregate a much greater number of benchmarks across wider range of difficulties than previous attempts to apply IRT to language model benchmarking (\cite{polo2024tinybenchmarksevaluatingllmsfewer, li2025adaptivetestingllmevaluation, truong2025reliableefficientamortizedmodelbased, dualindicators, zhou2025lostbenchmarksrethinkinglarge}).
This allows us to obtain a much longer time series of capability scores, which unlocks the novel applications we present in this work, such as measuring compute efficiency gains, detecting accelerations, and forecasting capabilities progress.

\section{Data}
\label{appendix:data}

\subsection{Benchmarks}
All of the benchmarks analyzed in this paper are taken from \href{https://epoch.ai/data/ai-benchmarking-dashboard}{Epoch AI's benchmarking hub}. These benchmarks are in turn separated into two categories -- internal benchmarks that are evaluated by Epoch AI themselves, and external benchmarks where results are obtained from external sources.

In both cases, the benchmarking hub prioritizes benchmarks that capture economically valuable tasks, are not yet saturated, and are widely used by AI researchers and practitioners. Internal benchmarks are subject to the additional restriction that the benchmark is easy to run.

We show the full list of internal and external benchmarks in the tables below. We include information about the benchmark's release date, as well as the core metric that we use (e.g. because some benchmarks may have multiple different metrics). In general, we stick to the metrics that most resemble an accuracy out of 100\%, and that most representative of the benchmark's overall difficulty (e.g. average performance across tasks in different subcategories of the benchmark). 

\subsection{Models}
We focused on collecting data from models that are highly cited, and also on models that are not heavily optimized for a particular benchmark.\footnote{Note that we also collect some information about the model evaluation setup, such as whether the model was evaluated under a few-shot setting. In cases where there are multiple models with different evaluation settings, we take the max over these model performances.} For example, there are many different agent scaffolds for SWE-Bench verified, but we do not include these. 

This serves two purposes: first, this is a pragmatic choice that helps prioritize data collection, since models with specific scaffolds are unlikely to be evaluated on multiple benchmarks, which is less helpful for helping fit the model. Second, this helps prevents skewed results, where models that are highly optimized for specific benchmarks end up with an unrepresentative estimated capability. 

Overall, a little over half of the models in our dataset are considered ``notable" based on the criteria in Epoch AI's Notable Models Database. In particular, this means that the models satisfy and of the following criteria (\cite{epoch2022pcdtrends}): 
\begin{itemize}
    \item Highly cited (over 5000 citations);
    \item Large training cost (over \$1,000,000, measured in 2023 USD, or at least 1\% the cost of the most expensive model trained to date, whichever is greater);
    \item Significant use (over one million monthly active users);
    \item State of the art performance (typically on a recognized benchmark);
    \item An equivalent level of historical significance, miscellaneous notability equal to or greater than the thresholds above, identified at the discretion of Epoch staff.
\end{itemize}

\begin{table}[htbp]
    \centering
    \caption{Internal benchmarks}
    \label{tab:benchmarks}
    \begin{tabular}{@{} l c c c c @{}}
        \toprule
        \textbf{Benchmark} & \textbf{Release Date} & \textbf{Metric} & \textbf{Source} \\
        \midrule
        FrontierMath & 2025/02/28 & \% correct (pass@1) & \cite{glazer2024frontiermathbenchmarkevaluatingadvanced}  \\
        GPQA Diamond & 2023/11/20 & \% correct (pass@1) & \cite{rein2023gpqagraduatelevelgoogleproofqa} \\
        MATH Level 5 & 2021/03/05 & \% correct (pass@1) & \cite{hendrycks2021measuringmathematicalproblemsolving} \\
        OTIS Mock AIME 2024-2025 & 2024/12/19 & \% correct (pass@1) & \cite{chen2025mockaime} \\
        SWE-Bench Verified & 2024/08/13 & \% correct (pass@1) & \cite{openai2024swebenchverified} \\
        \bottomrule
    \end{tabular}
\end{table}

\begin{table}[htbp]
    \centering
    \caption{External benchmarks}
    \label{tab:benchmarks}
    \begin{tabular}{@{} l c c c c @{}}
        \toprule
        \textbf{Benchmark} & \textbf{Release Date} & \textbf{Metric} & \textbf{Source} \\
        \midrule
        Adversarial NLI & 2019/10/31 & \% correct & \cite{nie2020adversarialnlinewbenchmark} \\
        Aider polyglot & 2024/12/21 & \% correct & \cite{aider2024polyglot} \\
        ARC-AGI & 2019/11/05 & \% correct & \cite{chollet2019measureintelligence} \\
        ARC Challenge & 2018/03/14 & \% correct & \cite{clark2018thinksolvedquestionanswering} \\
        Balrog & 2024/11/20 & Average progress & \cite{paglieri2025balrogbenchmarkingagenticllm} \\
        BIG-Bench Hard & 2022/10/17 & \% correct & \cite{suzgun2022challengingbigbenchtaskschainofthought} \\
        CadEval & 2025/04/22 & Pass rate & \cite{patrick2025cadevaldashboard} \\
        CommonSenseQA 2 & 2022/01/14 & Accuracy & \cite{talmor2022commonsenseqa20exposinglimits} \\
        Cybench & 2024/08/15 & Unguided \% Solved & \cite{zhang2025cybenchframeworkevaluatingcybersecurity} \\
        DeepResearch Bench & 2025/06/13 & Overall citation accuracy & \cite{du2025deepresearchbenchcomprehensivebenchmark} \\
        Factorio learning environment & 2025/03/06 & Lab success \% & \cite{hopkins2025factoriolearningenvironment} \\
        Fiction.LiveBench & 2025/02/21 & 16k token score & \cite{FictionLiveBench2025} \\
        GeoBench & 2025/03/01 & ACW Country \% & \cite{ccmdiGeoBench} \\
        GSM8K & 2021/10/27 & \% correct & \cite{cobbe2021trainingverifierssolvemath} \\
        GSO-Bench & 2025/05/29 & Success rate & \cite{shetty2025gsochallengingsoftwareoptimization} \\
        HellaSwag & 2019/05/19 & Accuracy & \cite{zellers2019hellaswagmachinereallyfinish} \\
        LAMBADA & 2016/06/20 & Accuracy & \cite{paperno2016lambadadatasetwordprediction} \\
        Lech Mazur Writing & 2025/01/31 & Mean rubric score & \cite{lechmazur2025writing} \\
        LiveBench & 2024/06/27 & Global average & \cite{white2025livebenchchallengingcontaminationlimitedllm} \\
        MMLU & 2020/09/07 & Accuracy & \cite{hendrycks2021measuringmassivemultitasklanguage} \\
        OpenBookQA & 2018/09/08 & \% correct & \cite{mihaylov2018suitarmorconductelectricity} \\
        OSUniverse & 2025/05/06 & Weighted score & \cite{davydova2025osuniversebenchmarkmultimodalguinavigation} \\
        OSWorld & 2024/04/11 & Score & \cite{xie2024osworldbenchmarkingmultimodalagents} \\
        PIQA & 2019/11/26 & Accuracy & \cite{bisk2019piqareasoningphysicalcommonsense} \\
        ScienceQA & 2022/09/20 & \% correct & \cite{lu2022learnexplainmultimodalreasoning} \\
        SimpleBench & 2024/10/31 & \% correct & \cite{simplebench2024} \\
        SuperGLUE & 2019/05/02 & Mean accuracy & \cite{wang2020supergluestickierbenchmarkgeneralpurpose} \\
        Terminal Bench & 2025/05/19 & Mean accuracy & \cite{terminalbench2025} \\
        The Agent Company & 2024/12/18 & \% resolved & \cite{xu2025theagentcompanybenchmarkingllmagents} \\
        TriviaQA & 2017/05/09 & Exact match & \cite{joshi2017triviaqalargescaledistantly} \\
        VideoMME & 2024/05/31 & \% overall (no subtitles) & \cite{fu2025videommefirstevercomprehensiveevaluation} \\
        Visual Physics Comprehension Test & 2025/01/30 & \% correct & \cite{browerVPCT2025} \\
        WeirdML & 2025/01/16 & Mean accuracy & \cite{ihle2025weirdml} \\
        WinoGrande & 2019/07/24 & Accuracy & \cite{sakaguchi2019winograndeadversarialwinogradschema} \\
        \bottomrule
    \end{tabular}
\end{table}

\begin{figure}
    \centering
    \includegraphics[width=\linewidth]{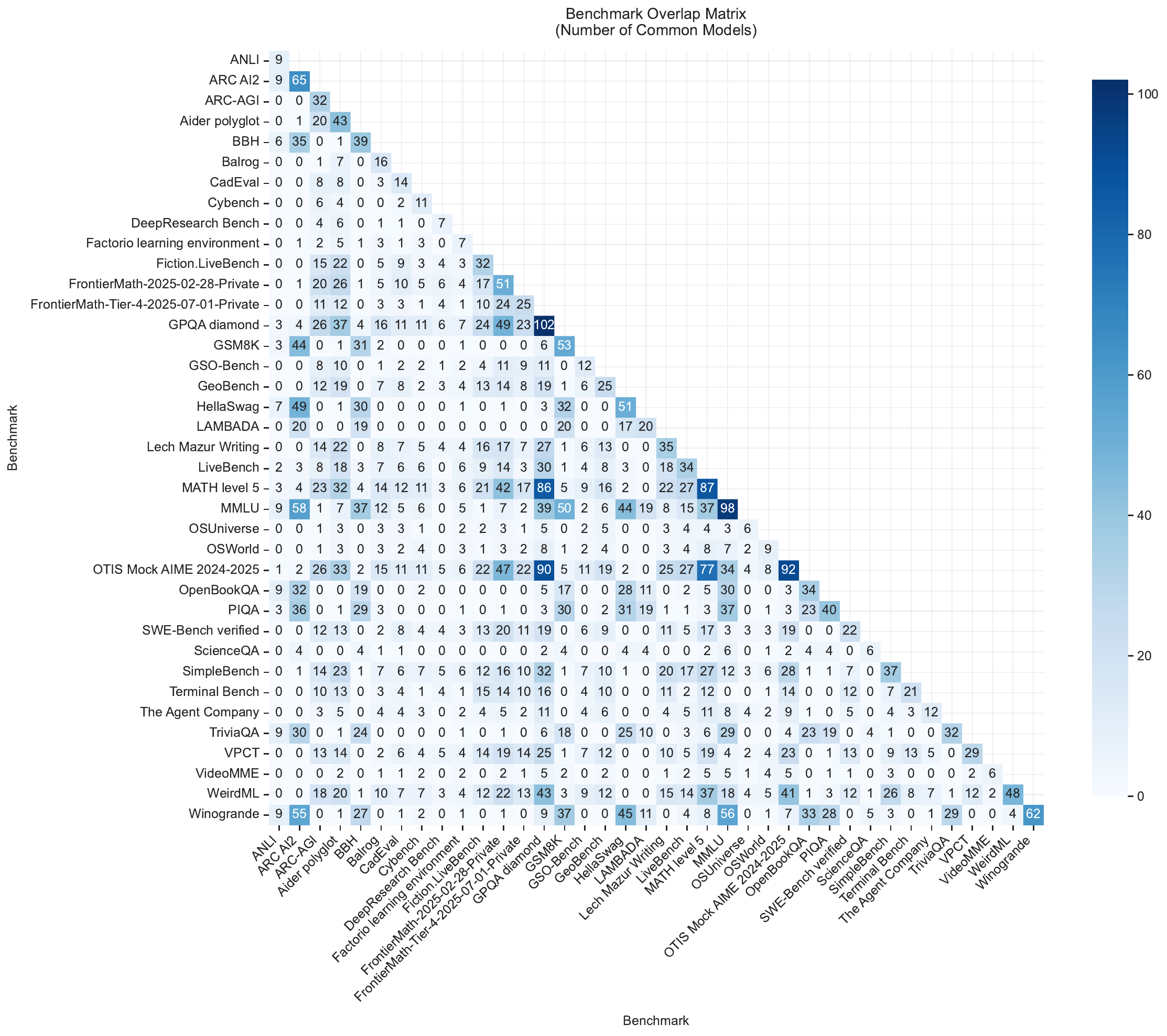}
    \caption{Overlap in the number of models that are evaluated on each of the benchmarks, as denoted by different benchmark ids. The benchmarks are sorted in chronological order (increasing date from left to right on the \(x\) axis).}
    \label{fig:overlap-matrix}
\end{figure}

\section{Algorithmic progress}
\label{appendix:algorithmic-progress}

Recall from Section \ref{sec:algorithmic-progress} that we estimated the rate of algorithmic progress by fitting the following model (a reiteration of Equation \ref{eq:algorithmic_progress}): 
\begin{equation*}
    C_m = k \log F_m + b.
\end{equation*}
The procedure is to first estimate \(k\) by looking at how estimated capabilities \(C_m\) vary with training compute \(F_m\) in specific families of models with the same training recipe. We then estimate \(b\) for each model using this \(k\), and looking at how \(b\) changes over time allows us to back out a rate of algorithmic progress. We performed this analysis on a dataset which does not include distilled models, to try and capture the relationship between compute and capabilities in frontier models. 

In this section we discuss two things: 
\begin{enumerate}
    \item Results using different modeling choices: Including distilled models in our dataset, estimating \(k\) without restricting to individual model families, looking at the frontier in algorithmic progress as opposed to capabilities, and potential algorithmic progress through variations in \(k\).
    \item An alternative approach for estimating algorithmic progress, e.g. by directly looking for examples of models that use less training compute at the same capability.
\end{enumerate}

\subsection{Alternative modeling choices}
\subsubsection{Including distilled models}
For our primary algorithmic progress analysis, we removed distilled models from our dataset -- this is because we want to understand the relationship between a model's capabilities and the compute used in its final training run, especially for frontier models. Many distilled models in our dataset are likely to be versions of frontier models that are less capable but substantially smaller and much cheaper to serve. 

However there are several issues with this. One is that it's not always clear whether a model is distilled purely from public information, since AI labs often do not disclose this kind of information about a model's development process. Furthermore, it's possible that models that ordinarily would be considered ``frontier" models (in the sense of pushing the state-of-the-art in capabilities) were actually distilled to some extent, e.g. it's possible that OpenAI's o3 model was partially distilled from GPT-4.5. 

Another issue is that if we're interested in understanding the relationship between final run training compute and model capabilities, we might also need to account for a wide range of other source of compute, e.g. synthetic data (\cite{epoch2025threeissuesunderminingcomputebasedaipolicies}). But it's likely the case that a wide range of recent AI models (e.g. in the last year) were trained on synthetic data, and it's not clear to us how much was used (and what threshold to use to exclude such models). 

Given these ambiguities, we also perform our analysis without dropping distilled models.\footnote{Note that the full set of distilled models can be found in the associated \href{https://github.com/epoch-research/benchmark-stitching}{GitHub repository}.}

Doing so allows us to add one additional family of models when estimating \(k\), namely the Qwen 2.5 family, yielding an average of \(k = 0.1625\). The corresponding rate of algorithmic progress is around 5\(\times\) per year, with \(\Delta b = 0.3\) per year (95\% CI: 0.0 to 0.5 per year). There is thus substantial overlap between these results and the ones we obtained in the main analysis in Section \ref{sec:algorithmic-progress}.

The full results from all families are shown in Table \ref{tab:all_models_compute_efficiency}. 

\begin{table}[h]
\centering
\begin{tabular}{lcccccc}
\hline
\textbf{Model Family} & \(k\) & \(\Delta b\) [95\% \textbf{CI}] & \multicolumn{2}{c}{\textbf{Annual compute efficiency gain} [95\% \textbf{CI}]} \\
\cline{4-5}
& & & \textbf{Model capabilities} & \textbf{Algorithmic quality} \\
\hline
Qwen & 0.15 & 0.40 [0.32, 0.48] & 8\(\times\) [4, 20] & 10\(\times\) [9, 20] \\
LLaMA & 0.18 & 0.38 [0.26, 0.49] & 5\(\times\) [3, 10] & 9\(\times\) [5, 10] \\
LLaMA 2 & 0.20 & 0.29 [-0.05, 0.63] & 3\(\times\) [2, 7] & 4\(\times\) [1, 10] \\
LLaMA 3.1 & 0.12 & 0.42 [0.38, 0.45] & 20\(\times\) [8, 40] & 30\(\times\) [20, 40] \\
\hline
\end{tabular}
\caption{Parameter estimates and rates of algorithmic progress across model families, without restricting to models that are not distilled or use little synthetic data. We derive 95\% confidence intervals for these by looking at the slope of \(b\) vs \(t\), and looking at the associated t-statistic and standard error.}
\label{tab:all_models_compute_efficiency}
\end{table} 

\subsubsection{Estimating \(k\) using all models with training compute estimates}
Besides choosing whether or not to include distilled models in our analysis, another degree of freedom comes from how we estimate \(k\). In the main analysis we did this by looking at the relationship between model capabilities \(C_m\) and log training compute \(\log F_m\), specifically within a particular ``training recipe" -- so the same algorithms and data, on the same family of models. But this approach has a weakness, which is that there are only a handful of families for which data is available (3 families in the main analysis), and there are also a limited number of models within each family (e.g. on the order of 5 models per family). 

We thus consider an alternative way of estimating \(k\) that doesn't restrict to using the same training recipe --- instead we plot \(C_m\) against \(\log F_m\) for all models for which data is available and look at the associated slope to determine \(k\). 

The major issue with this approach is that it introduces a bias --- we're likely to estimate a higher value of \(k\) than under the family-based approach, since algorithmic improvements would make you need less compute to get to the same performance. We then correspondingly find lower estimates of algorithmic progress. 
But there is a major upside, which is that we have a lot more models to work with -- in particular, we can consider all models for which we have training compute estimates available.

Performing this analysis yields an estimate of \(k = 0.22\) (95\% CI: 0.19 to 0.24) and \(\Delta b = 0.27\) per year (95\% CI: 0.20 to 0.34 per year) when looking at the frontier of algorithmic quality. The corresponding rate of algorithmic progress is \(3.5\times\) per year, with a 95\% CI of 2.6 to 4.8\(\times\) per year.  

Similarly, if we perform this analysis at the frontier of model capabilities, we find \(\Delta b = 0.21\) per year (95\% CI: 0.02 to 0.41 per year) when looking at the frontier of algorithmic quality. The corresponding rate of algorithmic progress is \(2.7\times\) per year, with a 95\% CI of 1.1 to 6.5\(\times\) per year. 

\subsubsection{What happens if algorithmic progress occurs by changing \(k\)?}
\label{appendix:scale_dependence_algorithmic_progress}

Our core approach to estimating algorithmic progress in Section \ref{sec:algorithmic-progress} assumes that all progress occurs by changing \(b\), but it's possible that this instead occurs by changing \(k\).\footnote{Another important way in which this model might be wrong is that the relationship between model capabilities \(C_m\) and log training compute \(\log F_m\) might not be linear.}
This latter effect introduces some degree of scale dependence into the model -- the decrease in the training compute needed to get to a certain capability might be higher at some levels of capability (or training compute) compared to others.

\begin{figure}
    \centering
    \includegraphics[width=1\linewidth]{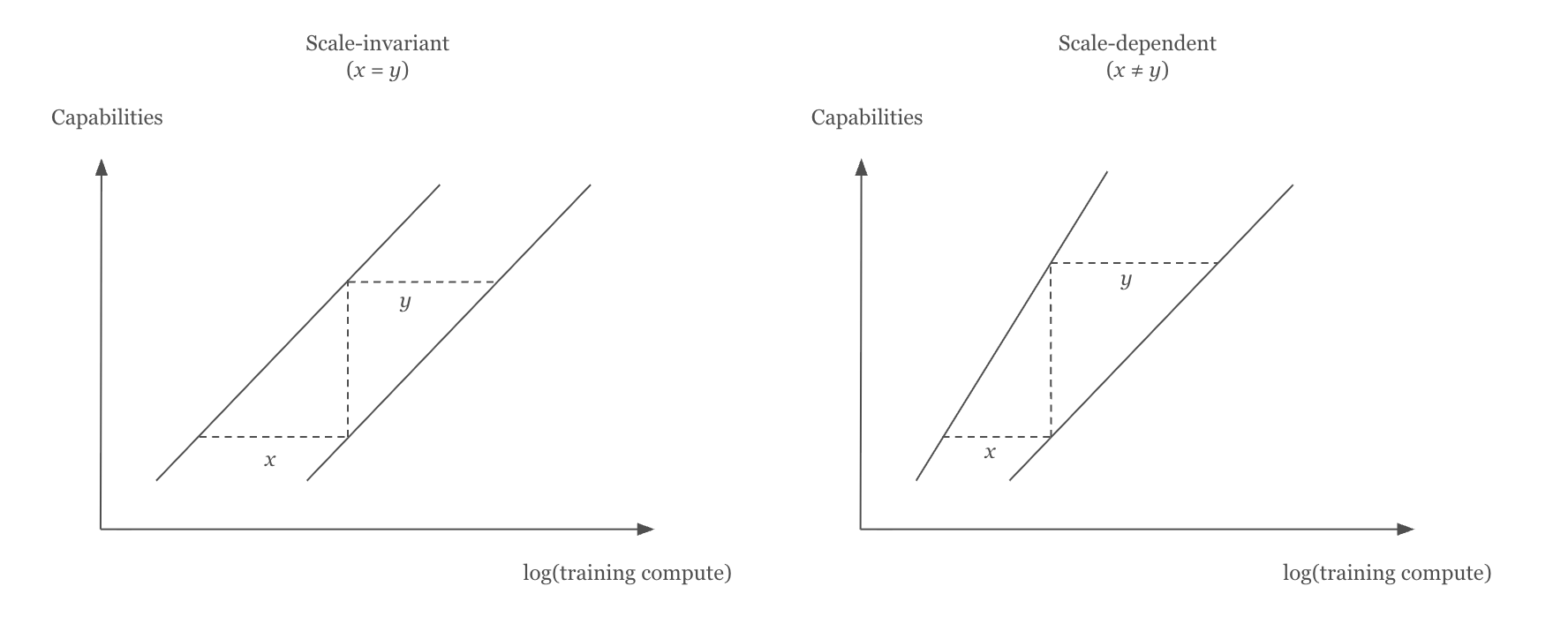}
    \caption{The observed rate of algorithmic progress might depend on the exact scale of compute \(F_m\) -- e.g. an algorithmic innovation that changes \(k\) might result in larger algorithmic efficiency gains at larger scales.}
    \label{fig:algorithmic_progress_scale_dependence}
\end{figure}

Another way of thinking about this is in terms of the increase in capabilities that can be attained at a fixed compute budget. This framing is especially important when it comes to questions about the ``software intelligence explosion" (\cite{eth2025willairdautomati, erdil2024estimating}), which involve massive capabilities improvements even without substantial increases in computational resources. In a world with scale-invariant algorithmic progress, the gain in capabilities at a fixed training compute budget is independent of the size of the training compute budget. In contrast, with scale-dependence it is possible that the gain in capabilities grows substantially as compute budgets grow. 

We illustrate this effect in Figure \ref{fig:algorithmic_progress_scale_dependence}. In the original model that only changes \(b\), a set of algorithmic innovations is seen as shifting the line relating capabilities \(C_m\) and log training compute \(\log F_m\) -- so the magnitude of algorithmic improvement is invariant with respect to the amount of training compute.


To be able to understand the magnitude of this effect we therefore want to get direct evidence about how much estimated capabilities increase at fixed training compute budgets, for compute budgets of different sizes. Unfortunately we do not have enough data to be able to attempt this (i.e. frontier models at substantially different training compute scales), so we leave this for future work.

\subsection{Directly estimating algorithmic progress}
\label{appendix:algorithmic_progress_direct_approach}

Rather than fitting the model based on Equation \ref{eq:algorithmic_progress}, we can also attempt to directly estimate the rate of algorithmic progress from individual data points. This gives us a way of sanity-checking our core results, although we consider these estimates less reliable overall --- hence we place them in the appendix rather than in the main paper. 

We then consider two ways of thinking about algorithmic progress, discussed in Appendix \ref{appendix:scale_dependence_algorithmic_progress}: (1) decreases in the training compute needed to achieve a certain level of estimated capability, and (2) increases in the estimated capability over time with a given budget of training compute. 

\textbf{Decreasing training compute needed to achieve a fixed capability}. 
For this approach, we first consider buckets with a narrow range of model capabilities. For example, this might be 0.3 capability units wide, going from 0.8 to 1.1. We then identify the first model in our data which has training compute data, and which falls into this bucket. Then we consider the set of models released after this, and trained with less compute, while also falling into the same bucket. Given this set, we then consider the models that push the frontier in algorithmic quality --- they use less compute than any previous model in the set. As long as there are three models that satisfy this criterion, we then fit a line of best fit, which tells us how much less training compute we need to get to the same capability. 

We sweep the buckets over the range of capabilities in our data, and show the resulting estimated annual compute reductions in Table \ref{tab:compute_efficiency_buckets_approach}. Like the estimates using our primary method in Section \ref{sec:algorithmic-progress}, the range of values is very wide. In particular, we find training compute reductions from 2\(\times\) to 400\(\times\)! The median estimate across these is around 10\(\times\) per year, but unfortunately we do not have much data and consider this method quite unreliable. 

\begin{table}[h]
\centering
\begin{tabular}{cccc}
\hline
Lower end of bucket & Upper end of bucket & Number of models & Annual compute reduction \\
\hline
0.8 & 1.1 & 12 & 5 \\
0.9 & 1.2 & 11 & 3 \\
1.2 & 1.5 & 9 & 400 \\
1.3 & 1.6 & 7 & 300 \\
1.4 & 1.7 & 12 & 10 \\
1.7 & 2.0 & 10 & 40 \\
1.8 & 2.1 & 8 & 20 \\
1.9 & 2.2 & 9 & 10 \\
2.0 & 2.3 & 8 & 2 \\
2.1 & 2.4 & 6 & 2 \\
\hline
\end{tabular}
\caption{We sanity check our algorithmic progress estimates in Section \ref{sec:algorithmic-progress} by direct observation, identifying models that match previous capabilities using less compute.}
\label{tab:compute_efficiency_buckets_approach}
\end{table}

\textbf{Increases in estimated capabilities at fixed training compute budgets}. 
Similar to the previous kind of algorithmic progress, we consider models that were initially at the frontier of capabilities. The difference is that we now consider the set of all models released at later dates that used about the same amount of training compute. In particular, given some reference compute level, we consider models that are \(\sqrt{2}\) times more or less compute intensive. Within this set of models, we then consider the frontier --- if there are at least 3 data points, we fit a trend line look at its slope. 

We show the results of this in Table \ref{tab:capability_gains_direct_approach}. Once again, the range of estimates is very wide, ranging from 0.1 to 1.6 capability units per year. The median increase is around 0.4 capability units per year. 

\begin{table}[h]
\centering
\begin{tabular}{ccc}
\hline
Bucket center (log10 FLOP) & Number of models & Capability units per year \\
\hline
23.05 & 7 & 1.6 \\
23.15 & 6 & 0.9 \\
23.25 & 7 & 0.4 \\
23.55 & 6 & 0.1 \\
23.75 & 6 & 0.3 \\
23.85 & 6 & 0.3 \\
23.95 & 7 & 0.3 \\
24.05 & 4 & 0.3 \\
24.45 & 8 & 0.4 \\
24.55 & 10 & 0.5 \\
24.65 & 9 & 0.4 \\
24.75 & 9 & 0.8 \\
24.85 & 6 & 0.7 \\
24.95 & 6 & 0.8 \\
25.25 & 5 & 0.4 \\
25.35 & 7 & 0.3 \\
25.45 & 9 & 0.2 \\
26.55 & 3 & 1.2 \\
26.65 & 3 & 1.2 \\
\hline
\end{tabular}
\caption{Estimated capability increases at fixed training compute budgets, by directly observing algorithmic improvements in the data.}
\label{tab:capability_gains_direct_approach}
\end{table}

\begin{figure}[htbp]
    \centering
    \begin{subfigure}[t]{0.48\linewidth}
        \centering
        \includegraphics[width=\linewidth]{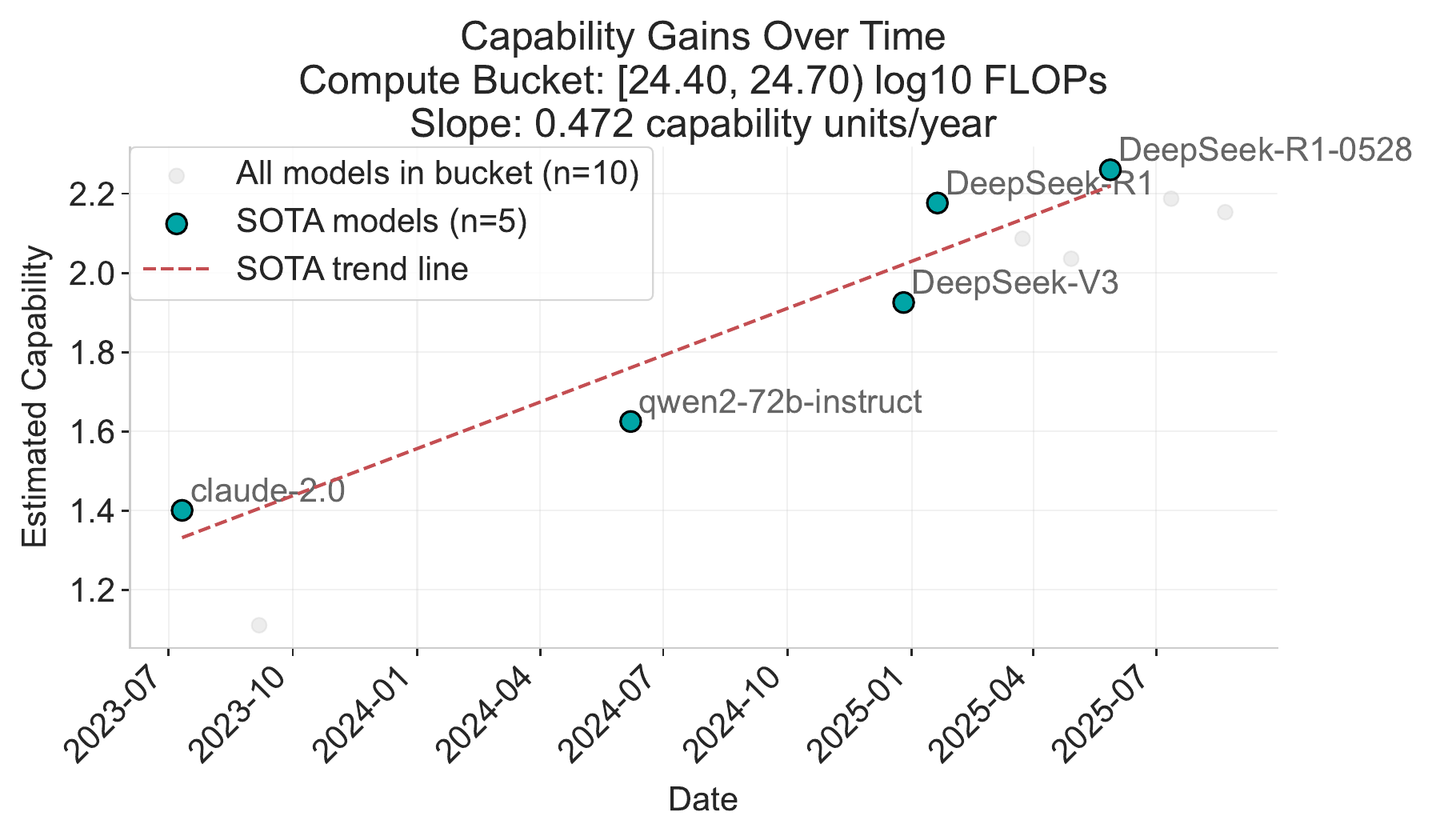}
        \caption{Increase in the estimated capabilities over time, at the same training compute budget.}
        \label{fig:capabilities-vs-time}
    \end{subfigure}
    \hfill
    \begin{subfigure}[t]{0.48\linewidth}
        \centering
        \includegraphics[width=\linewidth]{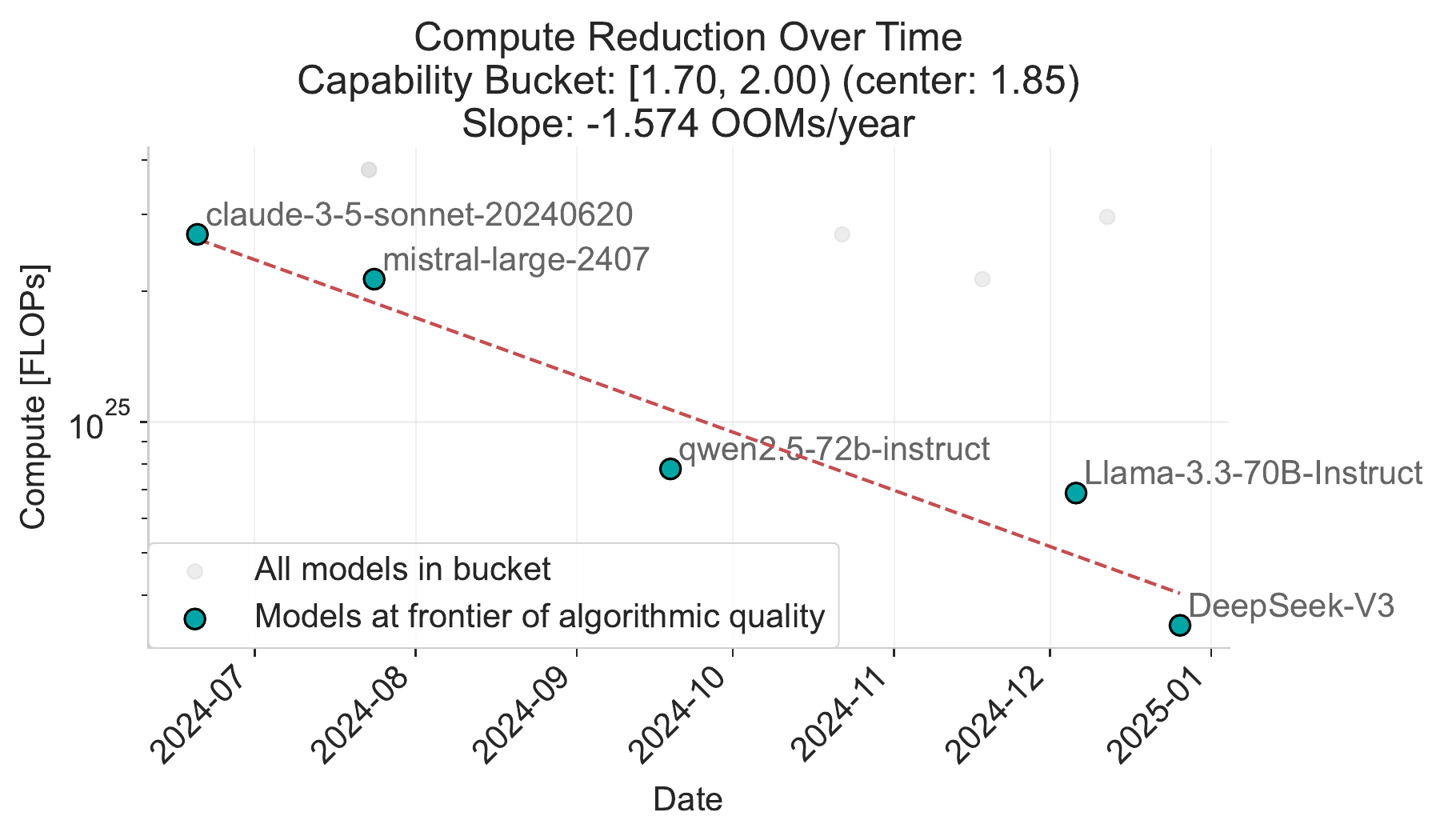}
        \caption{Reductions in the training compute required to reach a certain estimated capabilities level over time.}
        \label{fig:compute-efficiency-vs-time}
    \end{subfigure}

    \caption{Side-by-side comparison of predicted model capabilities and compute-efficiency trends.}
    \label{fig:capabilities-and-efficiency}
\end{figure}

We illustrate examples how we estimate both kinds of algorithmic progress directly from data points in Figure \ref{fig:capabilities-and-efficiency}. Overall, while this has merits on paper, it's especially sensitive to data limitations.\footnote{There may also be nuances in how to interpret ``compute efficiency improvement". For example, GPT-5 was released about a month after Grok 4 and used around \(5\times\) less training compute, while achieving similar performance. If we're not careful, the approach used in this section could pick this up as an algorithmic improvement of \(5^{12}\times\) per year!} To make the estimates reliable, we need to impose a range of constraints (e.g. a restricted capability range, or starting with frontier models) that reduce the number of data points that we can work with.\footnote{This also means that we do not have enough data on starting models that spanned multiple orders of magnitude of training compute, and thus cannot analyze the scale dependence of algorithmic progress (see Section \ref{appendix:scale_dependence_algorithmic_progress} for more detail on this).}

That being said, we overall have slightly more confidence in our estimates in Section \ref{sec:algorithmic-progress}, given that we obtain numbers that are broadly consistent with them here.

\subsection{Estimating algorithmic progress using ordinary least squares}
As a final approach to estimating algorithmic progress, we also consider a simple least squares regression. The model is specified in the following way: \begin{equation}
    C_m = k \log F_m + b (Y - Y_0),
\end{equation}
where \(Y\) is the model release date, and \(Y_0\) is the earliest model release date in the dataset. 

To determine the increase in model capabilities at fixed training compute, we can look directly at the estimate of \(b\), just as we did for the model in Section \ref{sec:algorithmic-progress}. To work out the compute efficiency improvement at the same capabilities, we consider \(k \log F_{m1} + b_1 = k\log F_{m2} + b_2\). This suggests that \(\frac{F_{m2}}{F_{m1}} = \exp\left(\frac{b_1 - b_2}{k}\right)\). 

This suggests a compute efficiency gain of around 20\(\times\) per year (95\% CI: 7 to 40) times. We also find an estimate of \(b\) of 0.37, with a 95\% CI of 0.29 to 0.46. 

The problem with this approach is that since algorithmic improvements are likely quite heavily correlated with compute scaling, it's hard to attribute which contributed more to increases in estimated capability. However, given that these estimates are broadly quite similar to our main approach (which does attempt to explicitly account for this) and the approach used in Appendix \ref{appendix:algorithmic_progress_direct_approach}, we think that the estimates likely have some merit. And we also find it useful for sanity-checking the results we obtained using other methods. 

\section{Synthetic data generation}
\label{appendix:synthetic-data}

In Section \ref{sec:detecting-acceleration}, we outlined synthetic data experiments illustrating how our approach can be used to detect future AI capability accelerations. Here we provide more detail about how this synthetic data was generated in the first place. 

We first generate synthetic data corresponding to a trend in model capabilities and benchmark difficulties over time, based on the rate of improvement around 0.55 units/year that we observed in Section \ref{sec:trends}. The model capabilities trend is piecewise linear with a single breakpoint (e.g. at the start of 2027), after which there is an acceleration by a certain factor \(N\). Benchmark difficulties have the same initial rate of increase but do not exhibit an acceleration. 

These are generated by specifying several parameters. The timeframe of consideration (e.g. 2020 to 2030) and the total capabilities increase (e.g. 3.5) specify the slope per year \(\Delta C\). We also specify a base capability \(C_{m0}\) at the start of the timeframe (e.g. 0), a model capabilities acceleration factor \(N\) (e.g. 2\(\times\)), and a noise term \(\epsilon \sim N(0, \sigma^2)\) (e.g. \(\sigma = 0.05\)).  
\begin{equation}
    C_m = \begin{cases}
        C_{m0} + t \Delta C + \epsilon & t \leq t_\text{cutoff} \\
        C_{m0} + t_\text{cutoff} \Delta C + (t - t_\text{cutoff}) N \Delta C & t \geq t_\text{cutoff}
    \end{cases}
    \label{eq:capabilities-synthetic-data}
\end{equation}

When generating specific datapoints, we sample uniformly from the time period (e.g. with 600 models) and determine the associated model capabilities based on Equation \ref{eq:capabilities-synthetic-data}. We also include Gaussian noise to make the synthetic data more realistic. 

We use a similar approach for generating benchmark difficulties \(D_m\), with a baseline difficulty \(D_{m0}\) (e.g. 0.5), slope \(\Delta D = \Delta C\), and noise \(\epsilon \sim N(0, \sigma^2)\). 
\begin{equation}
    D_m = D_{m0} + t \Delta D + \epsilon
    \label{eq:difficulties-synthetic-data}
\end{equation}
We choose 30 benchmarks to consider over the period of interest. 

What distinguishes the broad and narrow acceleration scenarios that we outlined in Section \ref{sec:trends} is the fraction \(f\) of the models that undergo the acceleration. In our main analysis, we considered a narrow acceleration scenario with \(f = 10\%\), and a broad acceleration scenario with \(f = 100\%\). 

We then map these model capabilities and benchmark difficulties to benchmark scores. For each model randomly select 25\% of all benchmarks to ``evaluate" on, and only keep performances that are within certain bounds (e.g. performance must be greater than \(\alpha = 5\%\) and less than \(1-\alpha = 95\%\)). 

\begin{figure}
    \centering
    \includegraphics[width=\linewidth]{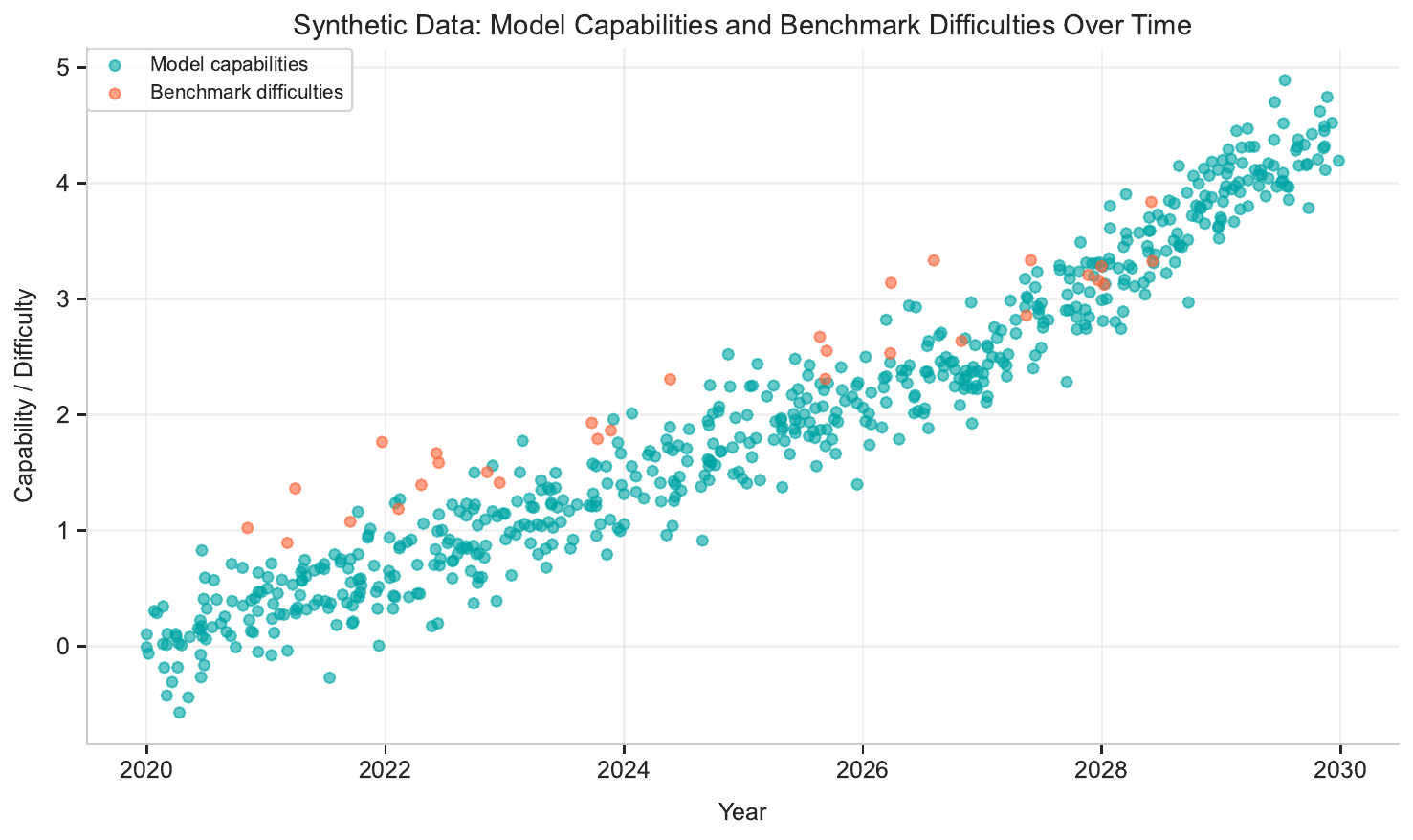}
    \caption{Synthetic data for the model capabilities and benchmark difficulties over time. Here we show an example where model capabilities accelerate two-fold in 2027, whereas benchmark difficulties improve linearly.}
    \label{fig:capabilities-difficulties-trend-broad}
\end{figure}

We show an example of the generated synthetic data in Figure \ref{fig:capabilities-difficulties-trend-broad}. 

\subsection{Noise structure}
In generating the synthetic data of model capabilities over time, we used a Gaussian noise structure. This is mainly for simplicity, and since our existing data has residuals that are approximately Gaussian. 

\begin{figure}[H]
    \centering
    \includegraphics[width=\linewidth]{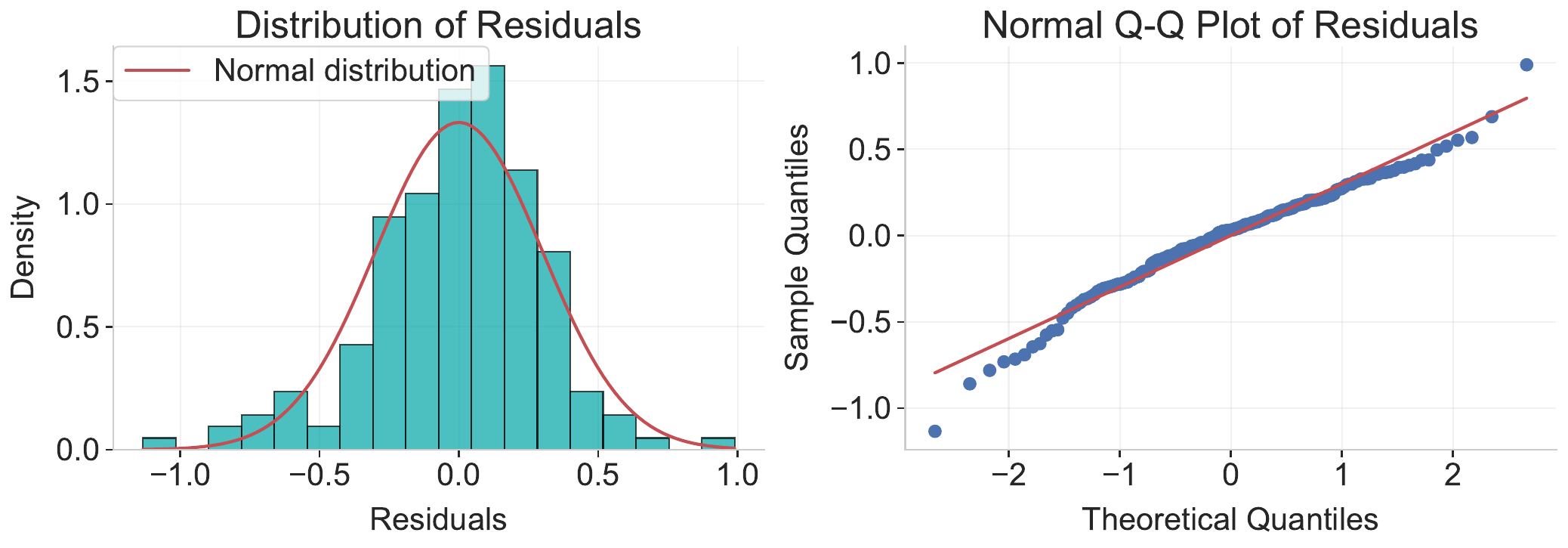}
    \caption{Fitting the data yields residuals that are approximately Gaussian, justifying the use of Gaussian noise in the synthetic data generation.}
    \label{fig:residuals}
\end{figure}

We can see this by plotting the residuals (see Figure \ref{fig:residuals}), as well as more rigorous statistical tests -- e.g. a Kolmogorov-Smirnov test is unable to reject that null hypothesis that the residuals are normally distributed (since the p-value is \(\approx 0.81\)). However, additional statistical tests (e.g. the Shapiro-Wilk test, Anderson-Darling test, and Jarque-Bera test) lead to the opposite conclusion at the same significance level. 

This is not critical for our use case since in this investigation we are primarily interested in the trend in frontier models. However, assuming a Gaussian noise structure remains a limitation, and a more meticulous analysis would model the noise structure more carefully.

\subsection{Detecting accelerations}
To test how much data is necessary for our model to detect accelerations, we sweep over the following parameters: 
\begin{itemize}
    \item \textbf{Models per year}: 40, 60, 80, 100, 120
    \item \textbf{Benchmarks per year}: 2, 4, 6, 8, 10
    \item \textbf{Post-cutoff acceleration}: 2.5, 5
\end{itemize}

\section{Robustness tests}
\label{appendix:robustness}

To check the robustness of our overall model fit, we consider making several variations. In this appendix we describe the results of our main robustness checks. In particular, we consider variations in which benchmarks are included in the statistical model fit, which statistical models are used, and also accounting for the evaluation setup. 

\subsection{Varying the amount of overlap}
As we alluded to in Section \ref{sec:data}, we filter out models that have only been evaluated on \(\leq 3\) of the benchmarks. This helps improve the model fit since it ensures high overlap between different benchmarks, helping calibrate the estimated benchmark difficulties and thus model capabilities. 

\begin{figure}[htbp]
    \centering
    \begin{subfigure}[b]{\textwidth}
        \includegraphics[width=\linewidth]{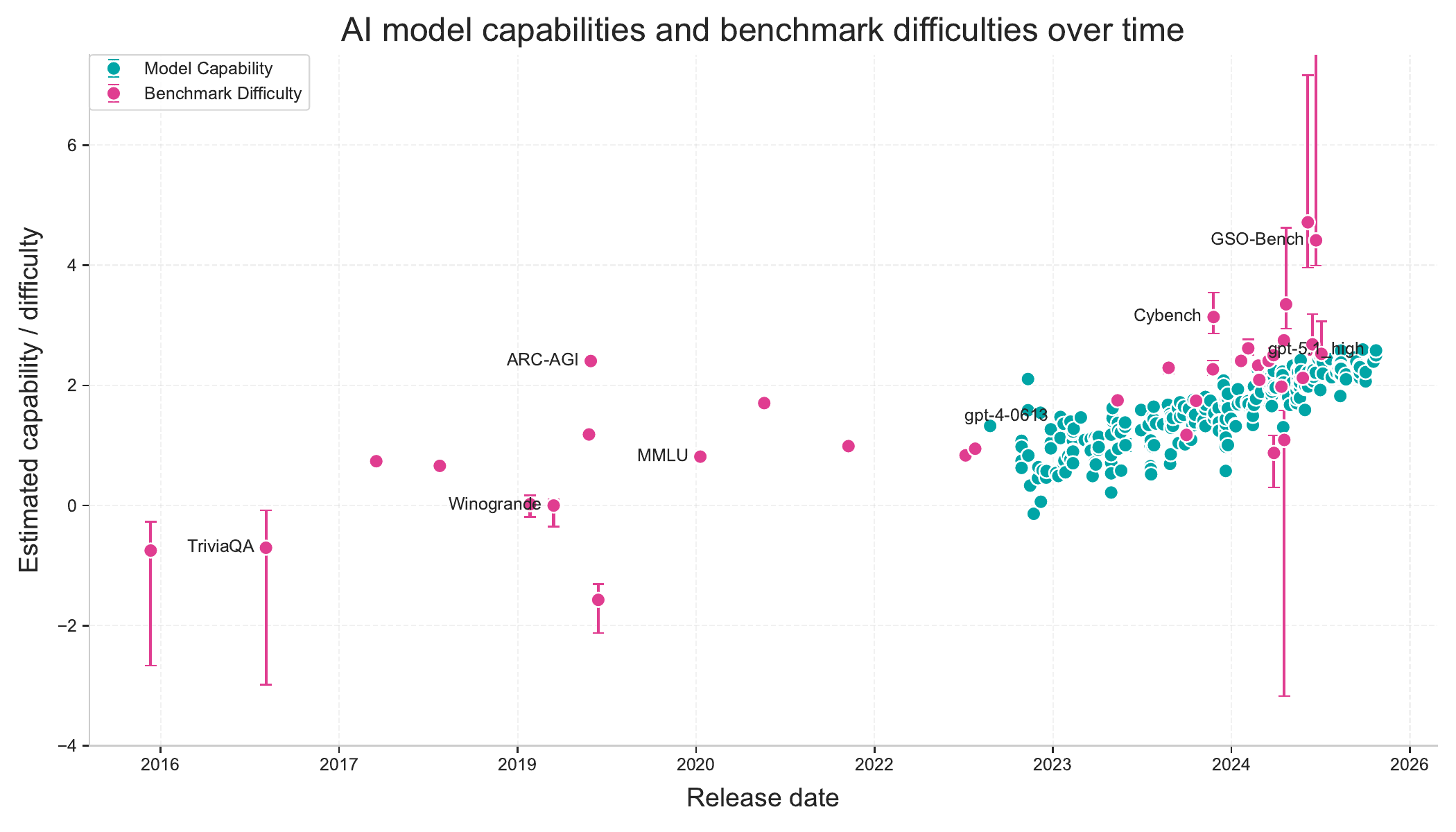}
        \caption{Fit when models are evaluated on at least 1 benchmark.}
        \label{fig:overlap-gt-0}
    \end{subfigure}
    \hfill
    \begin{subfigure}[b]{\textwidth}
        \includegraphics[width=\linewidth]{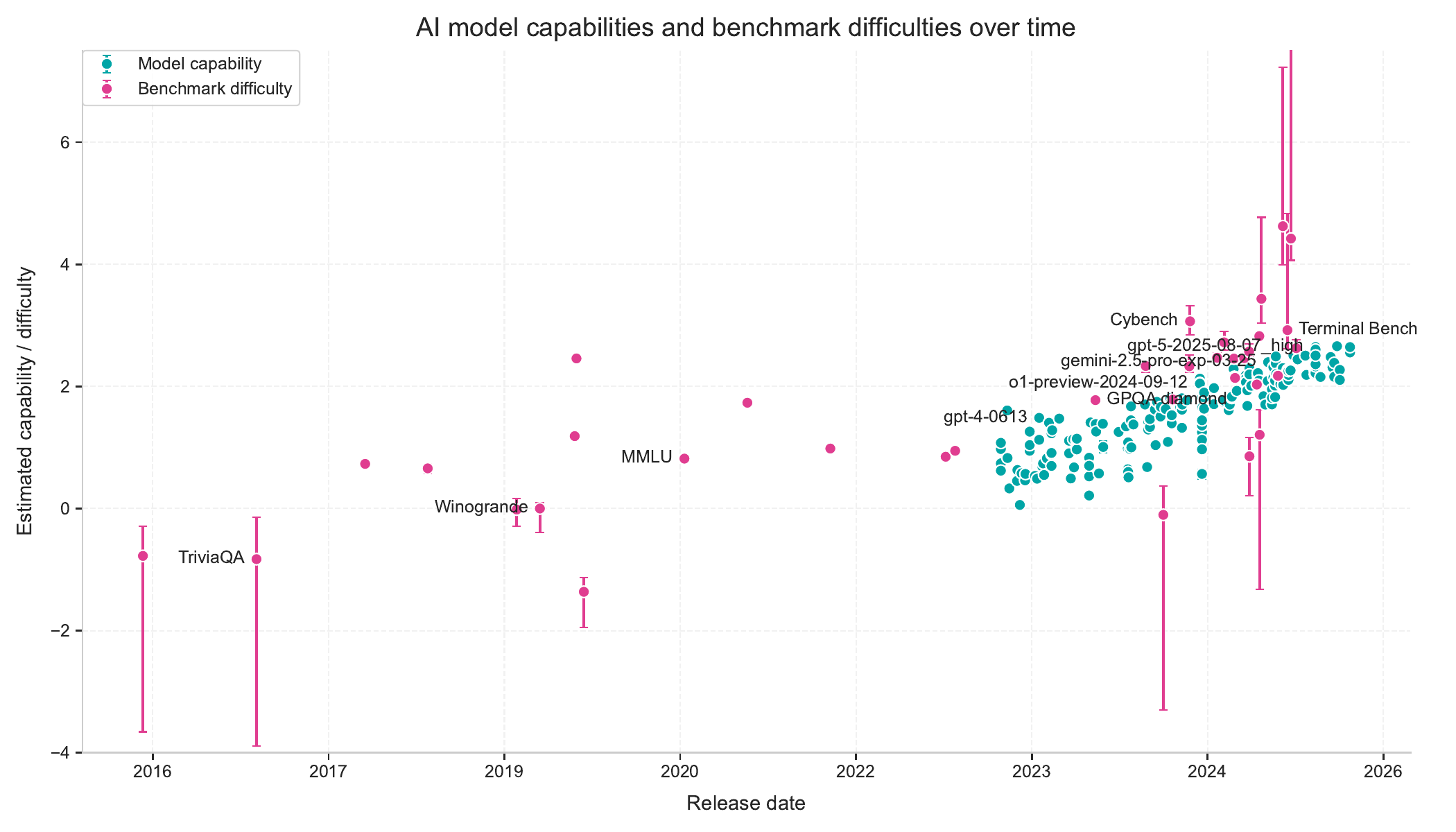}
        \caption{Fit when models are evaluated on at least 3 benchmarks.}
        \label{fig:overlap-gt-1}
    \end{subfigure}
    \caption{Model fits when varying the amount of overlap between different benchmarks, operationalized by filtering out models that are only evaluated below a certain number of benchmarks. Fits look very similar in both cases.}
    \label{fig:vary_overlap}
\end{figure}

If we vary the assumption we still end up with a similar set of estimated model capabilities, though decreasing the amount of required overlap increases the number of observed ``outliers" where models have substantially higher or lower estimated capabilities than expected. We show this in Figure \ref{fig:vary_overlap}.

\subsection{Varying the anchors}
In our main writeup, we maintain consistency in the absolute model capabilities and benchmark difficulties by removing two degrees of freedom from the model fit. In particular, we pick a single benchmark (WinoGrande), fix its difficulty to 0 and slope to 1. 

In this appendix, we attempt to vary this assumption by changing which benchmark is the anchor, and we also consider the possibility of using two separate model capabilities as the anchors instead of a single benchmark. 

\begin{figure}
    \centering
    \includegraphics[width=\linewidth]{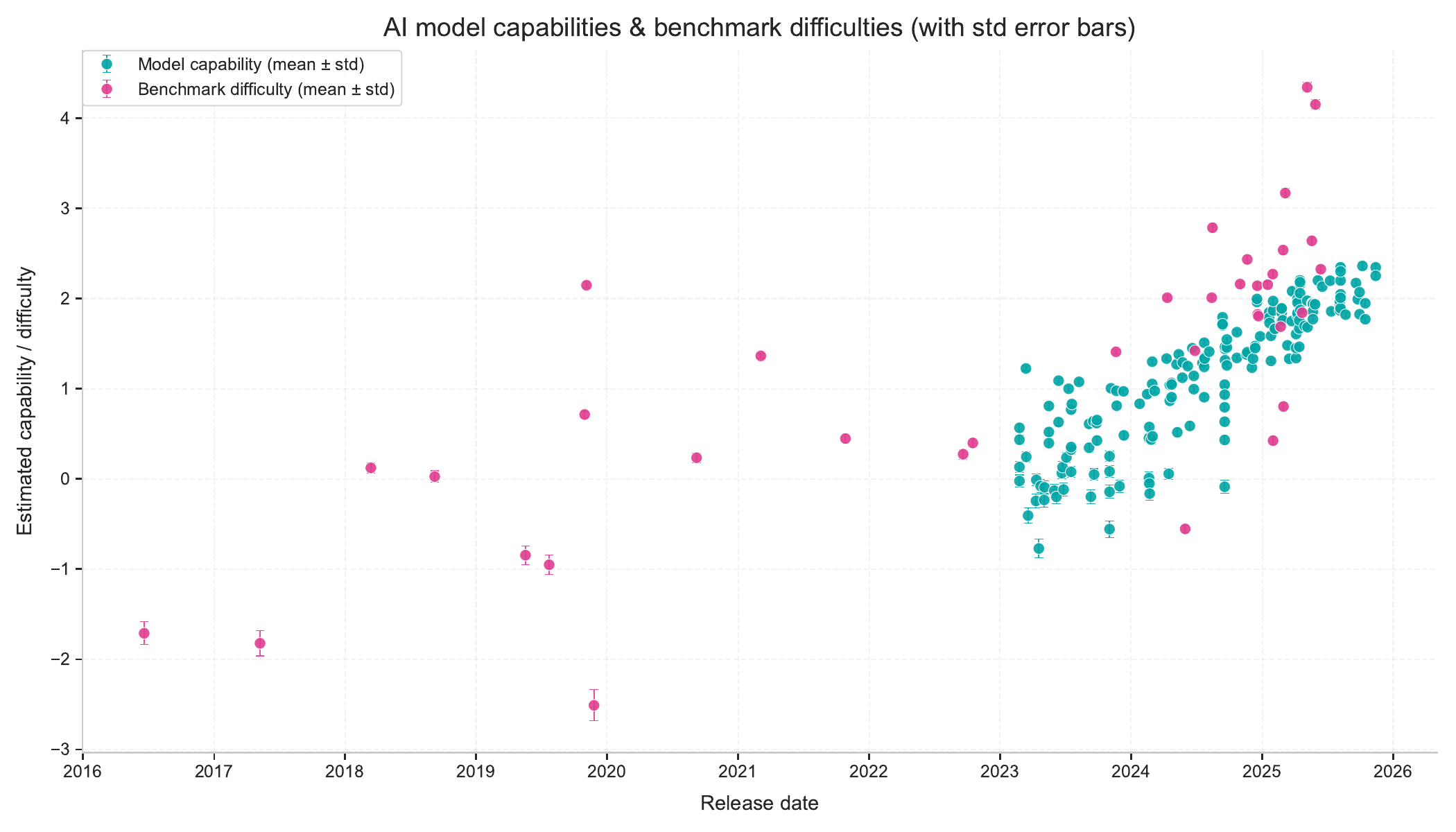}
    \caption{When changing benchmark anchors, the change in the estimated model capabilities and benchmark difficulties tends to be very small. Error bars show the range in estimated model capabilities and benchmark difficulties when varying the benchmark anchor.}
    \label{fig:change_anchors}
\end{figure}

When varying the benchmark anchor, we see that there is some degree of variation, and we plot the standard deviation of this as error bars around the main model estimates in Figure \ref{fig:change_anchors}.

\subsection{Varying benchmark inclusion}
As we allude to in Section \ref{sec:discussion}, the model fit is sensitive to the specific set of included benchmarks. In this section we test the robustness of our approach to this, considering different possible selections of the set of included benchmarks. 

We consider two separate ways of varying benchmark inclusion: (1) randomly selecting some subset of the overall set of benchmarks, and (2) benchmarks that are vs are not ``optimized-for"  by labs. 

\subsubsection{Random selection}
The first way we vary benchmark inclusion is to randomly drop some fraction of the benchmarks and compare the model fits in each case. We first fit the statistical model, and then estimate (1) the capability improvement per year, and (2) the total range of model capabilities without regard for time. We do this for 100 separate random selections, and find a median slope of 0.54 capability units per year, which is very similar to our central estimate of 0.55 in our main analysis. The 95\% confidence interval across these 100 random selections is 0.43 to 0.64.

\subsubsection{Detecting Benchmark Gaming}

A common concern about AI models is that they can be heavily optimized for particular benchmarks, or that benchmark scores could be selectively reported, leading to impressive benchmark results without a commensurate increase in general capabilities (e.g. see \cite{dominguezolmedo2025trainingtesttaskconfounds}). 

Our statistical framework provides a way of testing this - the idea is to compare the estimated capabilities when we fit the model on two subsets of the full dataset: (1) scores on benchmarks that have been ``optimized-for" by AI labs, and (2) scores on benchmarks have \emph{not} been ``optimized-for". We then check if the estimated capabilities are substantially higher on the optimized-for set compared to the not-optimized-for set.

For the purposes of this analysis, we define a benchmark as "optimized-for" if a frontier AI lab publicly reports their model's performance on that benchmark in at least one instance.\footnote{The AI labs that we (somewhat arbitrarily) consider are OpenAI, Anthropic, xAI, Google DeepMind, Meta, DeepSeek, Alibaba, and Mistral. We attempt to identify reported model performances from these labs by searching through their blog posts, papers, and videos. The full list of benchmarks and whether they have been "optimized-for" is listed in Table \ref{tab:benchmarks-optimized-for} of Appendix \ref{appendix:robustness}.}

Previously in subsection \ref{sec:optimization}, we used an anchor benchmark (e.g. WinoGrande) with a fixed difficulty slope to address identifiability issues. However, any benchmark we choose as anchor cannot appear in both the optimized-for and not-optimized-for sets, which prevents us from measuring the effect of optimization pressure directly. We therefore repeat the analysis with multiple anchor benchmarks from both sets.

Our procedure is as follows. First, we filter our data to select only benchmarks created in 2024 and later. This is necessary because the not-optimized-for benchmarks tend to be more recent, and therefore more difficult than the optimized-for benchmarks. Filtering ensures that both categories of benchmarks are of similar difficulty.

Next, we select the 9 optimized-for benchmarks with the highest model coverage (there were only 9 such benchmarks left after filtering for recency) and the 10 not-optimized-for benchmarks with the highest model coverage. For each of these 19 benchmarks, we perform two capability fits: one where it serves as anchor for its original set (optimized-for or not-optimized-for), and one where it serves as anchor after being moved to the opposite set. This gives us 40 capability estimates per model. For each anchor benchmark, we compute the mean difference in model capabilities between the optimized-for and not-optimized-for benchmark sets. 

We then average this difference across all 20 anchor benchmarks.
This analysis finds a small difference: averaging over the 20 anchor benchmarks, estimated model capabilities are only 0.12 capability units higher on the optimized-for set compared to the not-optimized-for set, i.e. a sixth of a year's worth of capabilities progress. 

To check  whether this effect is statistically significant, we perform a permutation test. For each anchor benchmark, we randomly partition the remaining benchmarks into two equally-sized sets (adding the anchor to both), fit model capabilities on each partition, and compute the capability difference between them. We repeat this process 100 times per anchor benchmark to generate a null distribution.

We find that the observed capability differences are not much larger than expected under random partitioning. Across the 20 anchor benchmarks, we find p-values ranging from 0.1 to 0.37, with an average value of 0.22. Therefore, our results do not provide statistically significant evidence of model developers optimizing for or strategically reporting benchmark scores.

\begin{figure}
    \centering
    \begin{subfigure}[b]{0.48\textwidth}
        \centering
        \includegraphics[width=\textwidth]{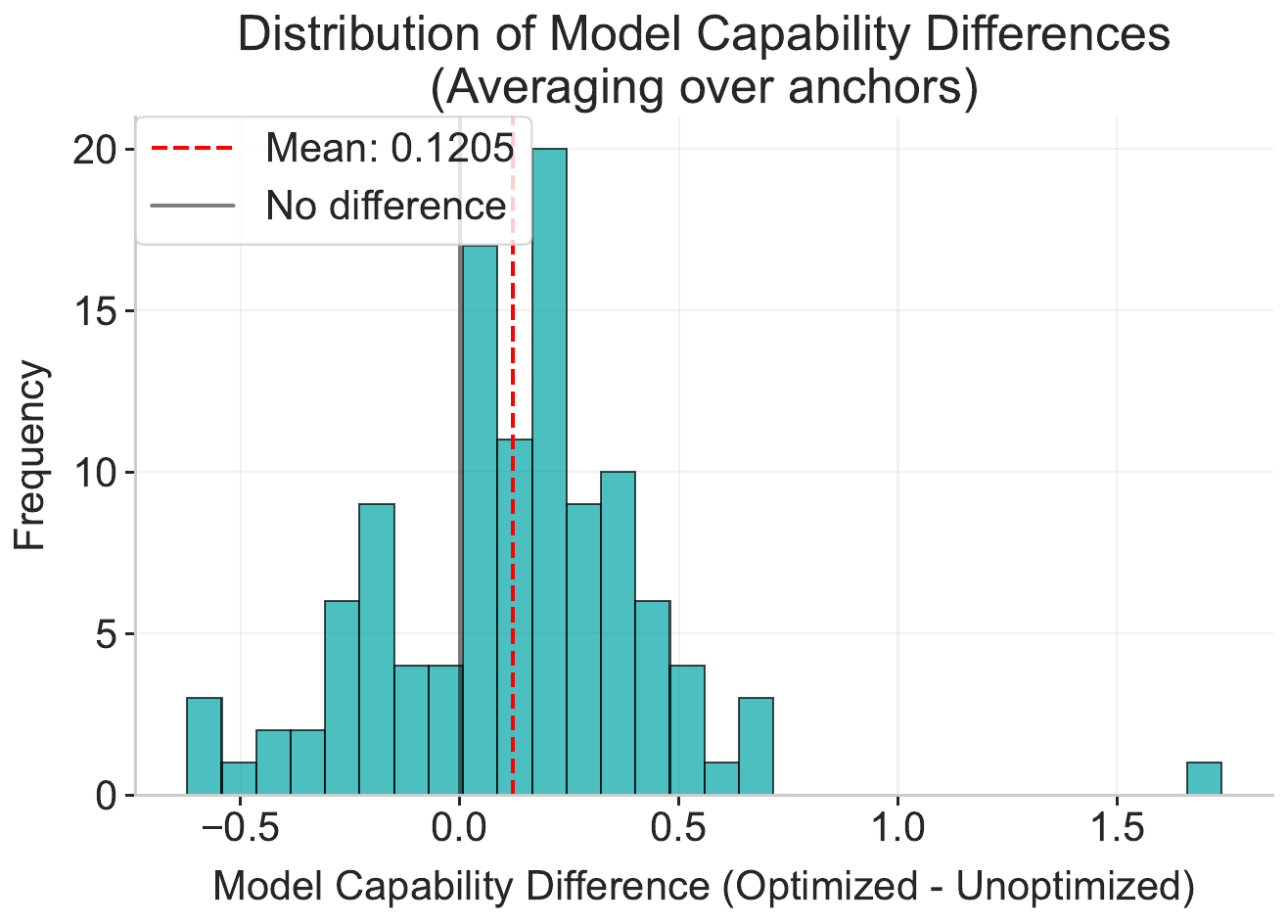}
        \label{fig:optimized_vs_unoptimized_histogram}
    \end{subfigure}
    \hfill
    \begin{subfigure}[b]{0.48\textwidth}
        \centering
        \includegraphics[width=\textwidth]{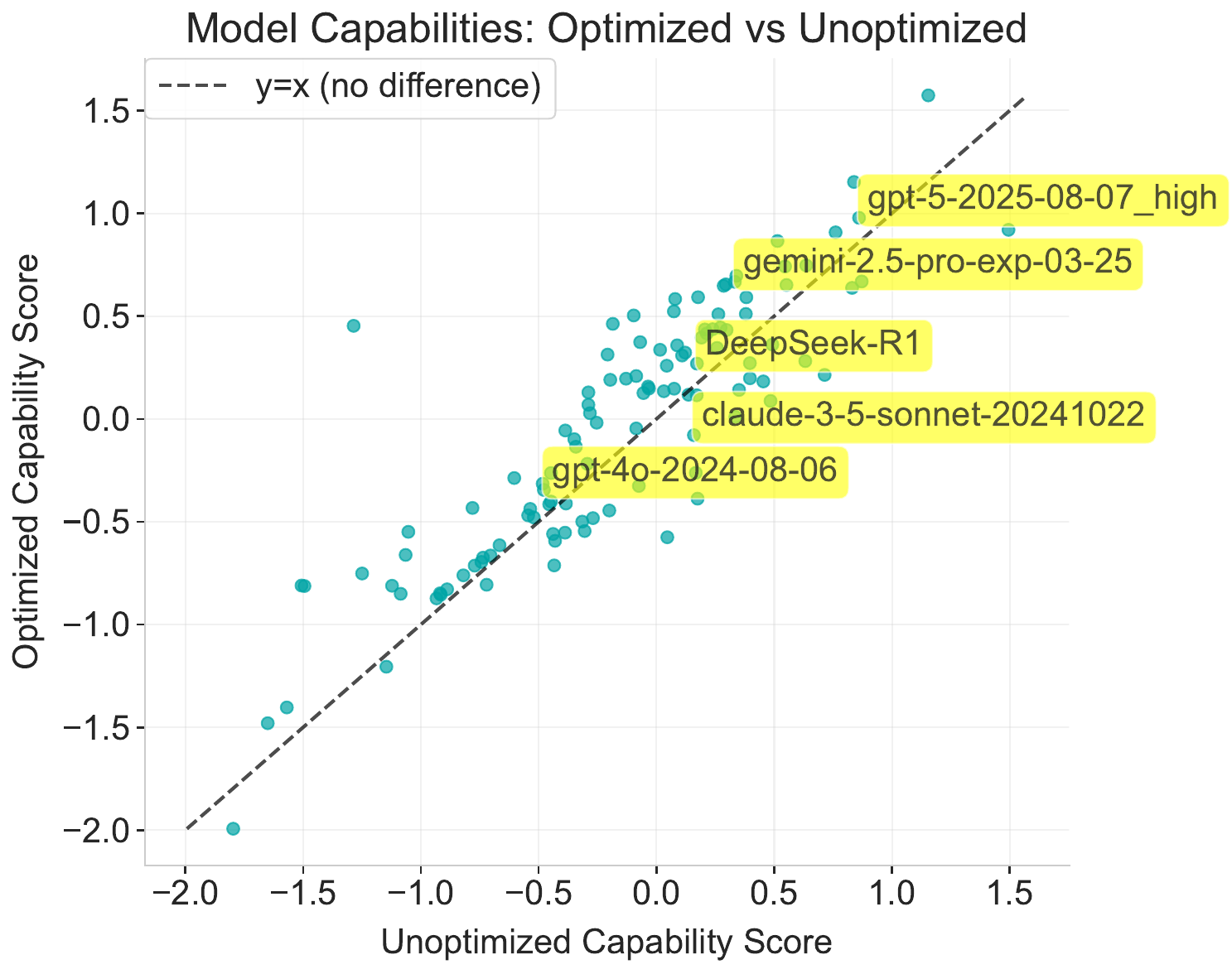}
        \label{fig:optimized_vs_unoptimized_scatter}
    \end{subfigure}
    \caption{Estimated capabilities when fitting on optimized-for benchmarks are slightly, but not statistically significantly higher than when fitting on not-optimized-for benchmarks.}
    \label{fig:benchmark_comparison}
\end{figure}

\begin{table}[htbp]
    \centering
    \caption{List of benchmarks and whether or not they have been optimized for. We consider a benchmark as ``optimized for" if it was reported in at least one frontier AI lab.}
    \label{tab:benchmarks-optimized-for}
    \begin{tabular}{l c}
      \textbf{Benchmark} & \textbf{Optimized for} \\ 
      \hline
      Aider polyglot                          & TRUE  \\
      ANLI                                    & TRUE  \\
      ARC AI2                                 & TRUE  \\
      ARC-AGI                                 & TRUE  \\
      Balrog                                  & FALSE \\
      BBH                                     & TRUE  \\
      BoolQ                                   & TRUE  \\
      CadEval                                 & FALSE \\
      CSQA2                                   & FALSE \\
      Cybench                                 & TRUE  \\
      DeepResearch Bench                      & FALSE \\
      Factorio learning environment           & FALSE \\
      Fiction.LiveBench                       & FALSE \\
      FrontierMath-2025-02-28-Private         & TRUE  \\
      GeoBench                                & FALSE \\
      GPQA diamond                            & TRUE  \\
      GSM8K                                   & TRUE  \\
      GSO-Bench                               & FALSE \\
      HellaSwag                               & TRUE  \\
      LAMBADA                                 & TRUE  \\
      Lech Mazur Writing                      & FALSE \\
      LiveBench                               & TRUE  \\
      MATH level 5                            & TRUE  \\
      MCBench                                 & FALSE \\
      MMLU                                    & TRUE  \\
      OpenBookQA                              & TRUE  \\
      OSUniverse                              & FALSE \\
      OSWorld                                 & TRUE  \\
      OTIS Mock AIME 2024-2025                & TRUE  \\
      PIQA                                    & TRUE  \\
      ScienceQA                               & TRUE  \\
      SimpleBench                             & FALSE \\
      SuperGLUE                               & TRUE  \\
      SWE-Bench verified                      & TRUE  \\
      Terminal Bench                          & TRUE  \\
      The Agent Company                       & FALSE \\
      TriviaQA                                & TRUE  \\
      VideoMME                                & TRUE  \\
      VPCT                                    & FALSE \\
      WeirdML                                 & FALSE \\
      Winogrande                              & TRUE  \\
    \end{tabular}
\end{table}

\subsection{Varying the statistical model}
\label{appendix:varying_statistical_model}
We also consider modifying Equation \ref{equation:main-model} such that the model fit is not a sigmoid. In particular, we also follow \cite{owen2024predictable} in considering a clipped linear function as well. In particular, this is specified as follows: 
\begin{equation}
    \text{performance}(m,b) = \text{clip}(\alpha_b (C_m - D_b), [0,1]).
    \label{eq:clipped-linear}
\end{equation}

To distinguish between these, we compare the fits from the two models. We find that both models perform similarly well in cross-validation. 

\begin{table}[ht]
  \centering
  \caption{Results from model fits and 10-fold cross validation.}
  \begin{tabular}{l r r r r r}
    \textbf{Model type} & \(R^2\) & \textbf{MSE} & \textbf{AIC} & \textbf{BIC} & \textbf{No. of params} \\
    \hline
    Sigmoid         & 0.8641 & 0.0098 & -6490.3797 & -5177.7113 & 253 \\
    Clipped linear & 0.8683 & 0.0095 & -6472.1491 & -5159.4807 & 253 \\
  \end{tabular}
  \label{tab:model_comparison}
\end{table}

\subsection{Analysis over a longer time period}
\label{sec:data_range_limitations}

\begin{figure}
    \centering
    \includegraphics[width=\linewidth]{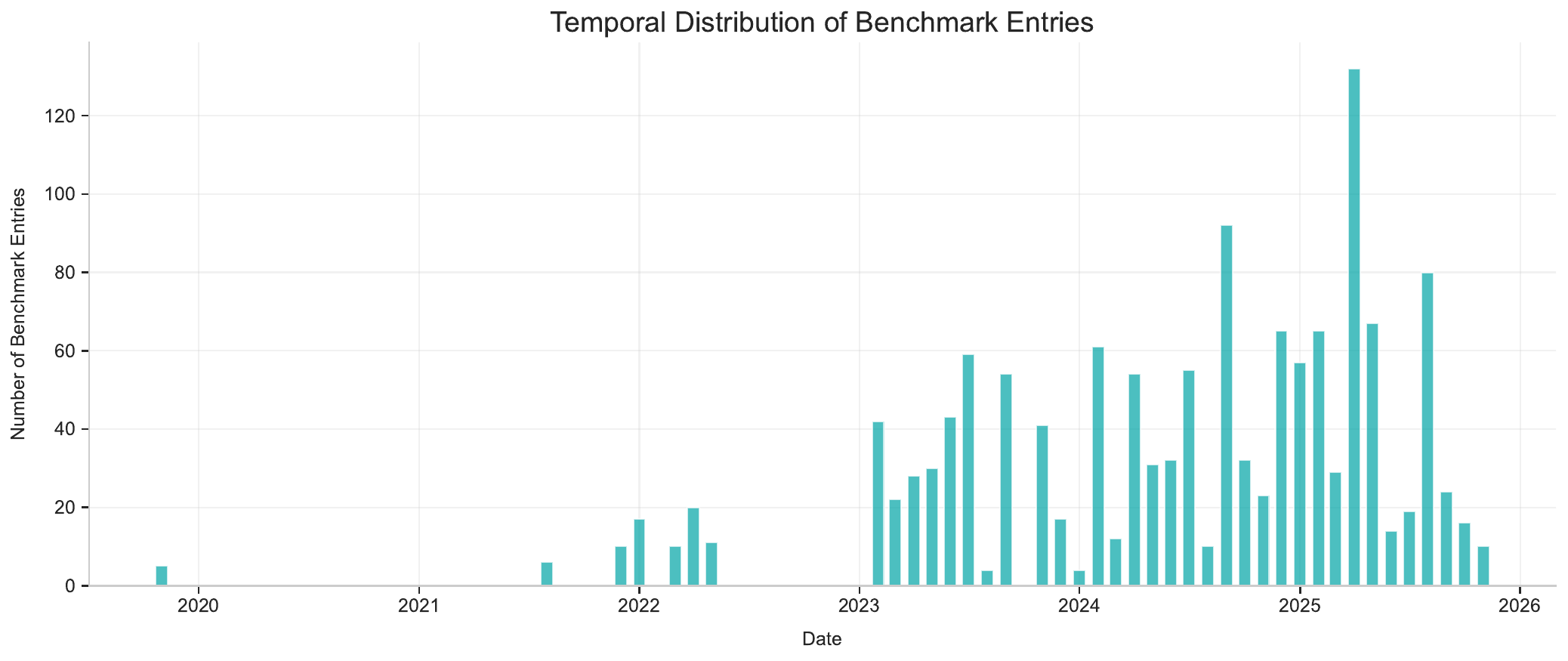}
    \caption{The vast majority of our data is from 2023 or later, and earlier results are substantially more unreliable.}
    \label{fig:data_over_time}
\end{figure}

In our main analysis in Section \ref{sec:results}, we only work with data from 2023 or later. The main reason for this is that our data is much sparser in earlier years, with only a handful of models between 2019 and 2023 (see Figure \ref{fig:data_over_time}) -- but having many models evaluated on benchmarks with substantial temporal overlap is important for our approach to give reliable results. These earlier models are also less representative for the purposes of forecasting future capabilities. 

However, this introduces an additional degree of freedom into the analysis that risks overfitting the data, so in this section we also share results without removing this earlier data (around 100 benchmark entries). Our findings are similar overall.

\subsubsection{Forecasting future capabilities}

\begin{figure}
    \centering
    \includegraphics[width=\linewidth]{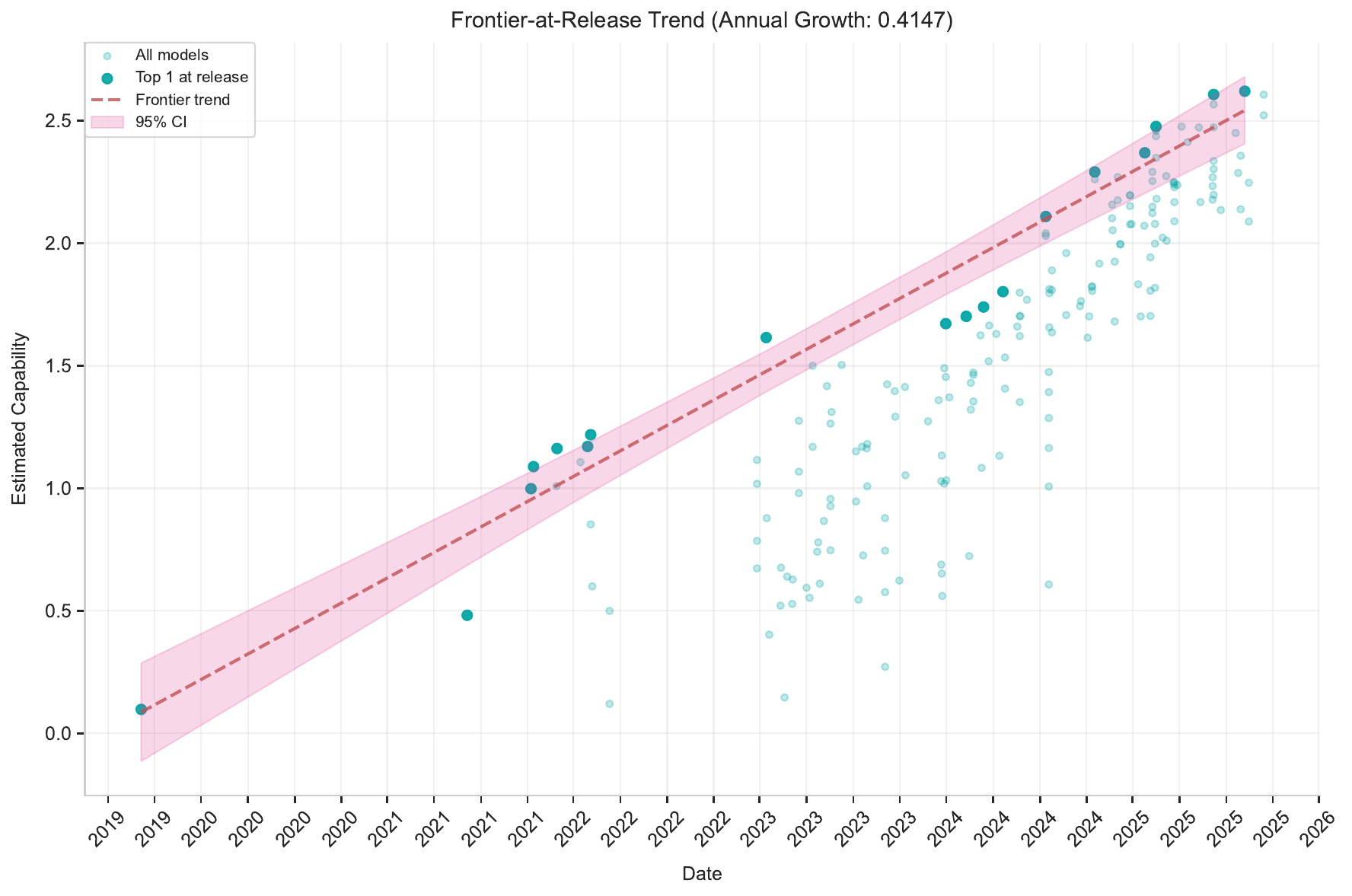}
    \caption{The vast majority of our data is from 2023 or later, and earlier results are substantially more unreliable.}
    \label{fig:frontier_at_release_trend}
\end{figure}

When we perform this analysis, we find a slower overall rate of progress over time, at around 0.42 capability units per year [95\% CI: 0.37 to 0.46]. Instead of 4.5 capability units in three years, this predicts around 3.8 capability units. And like with our main results, the forecasts made using this approach would have underestimated the rate of progress over the last year, and to an even greater degree. We show this in Figure \ref{fig:frontier_at_release_trend}. 

We also use our same acceleration detection pipeline, and find that there was a potential acceleration of around 1.7\(\times\) in mid-2024. This is broadly consistent with the analysis that we did with the data when restricted to post-2023. 

\textbf{GPT-3 to GPT-4 to GPT-5}
Many people were less impressed by the jump from GPT-4 to GPT-5 compared to the one from GPT-3 to GPT-4. But our statistical framework predicts that the jump from GPT-4-to-GPT-5 was possibly twice as large as the one from GPT-3-to-GPT-4! We show this in Table \ref{tab:gpt_capabilities}. 

Why is there a discrepancy between the model's predictions and how people felt about the sizes of these two jumps? One possible reason is that there were many more notable releases between GPT-4 and GPT-5, like GPT-4o and o3, compared to between GPT-3 and GPT-4. Another possible reason is that our model's predictions for some of these earlier models are especially unreliable, since they're based on sparser data. 

\begin{table}[h]
\centering
\begin{tabular}{lc}
\hline
\textbf{Model} & \textbf{Estimated Capability} \\
\hline
gpt-5-2025-08-07\_high & 2.61 \\
o3-2025-04-16\_high & 2.48 \\
o1-2024-12-17\_high & 2.26 \\
gpt-4.5-preview-2025-02-27 & 2.08 \\
gpt-4o-2024-11-20 & 1.76 \\
gpt-4-0314 & 1.62 \\
text-davinci-001 & 1.01 \\
\hline
\end{tabular}
\caption{Estimated capabilities for selected GPT and o-series models. Notably, the increase in capabilities when going from GPT-3 (text-davinci-001) to GPT-4 was around 0.6 units, whereas the jump from GPT-4 to GPT-5 with the highest reasoning setting was closer to 1 unit. So according to the model, the latter jump was a little less than twice as large as the former.}
\label{tab:gpt_capabilities}
\end{table}


\section{Model fit}
\label{appendix:model-fit}

\begin{figure}
    \centering
    \includegraphics[width=0.8\linewidth]{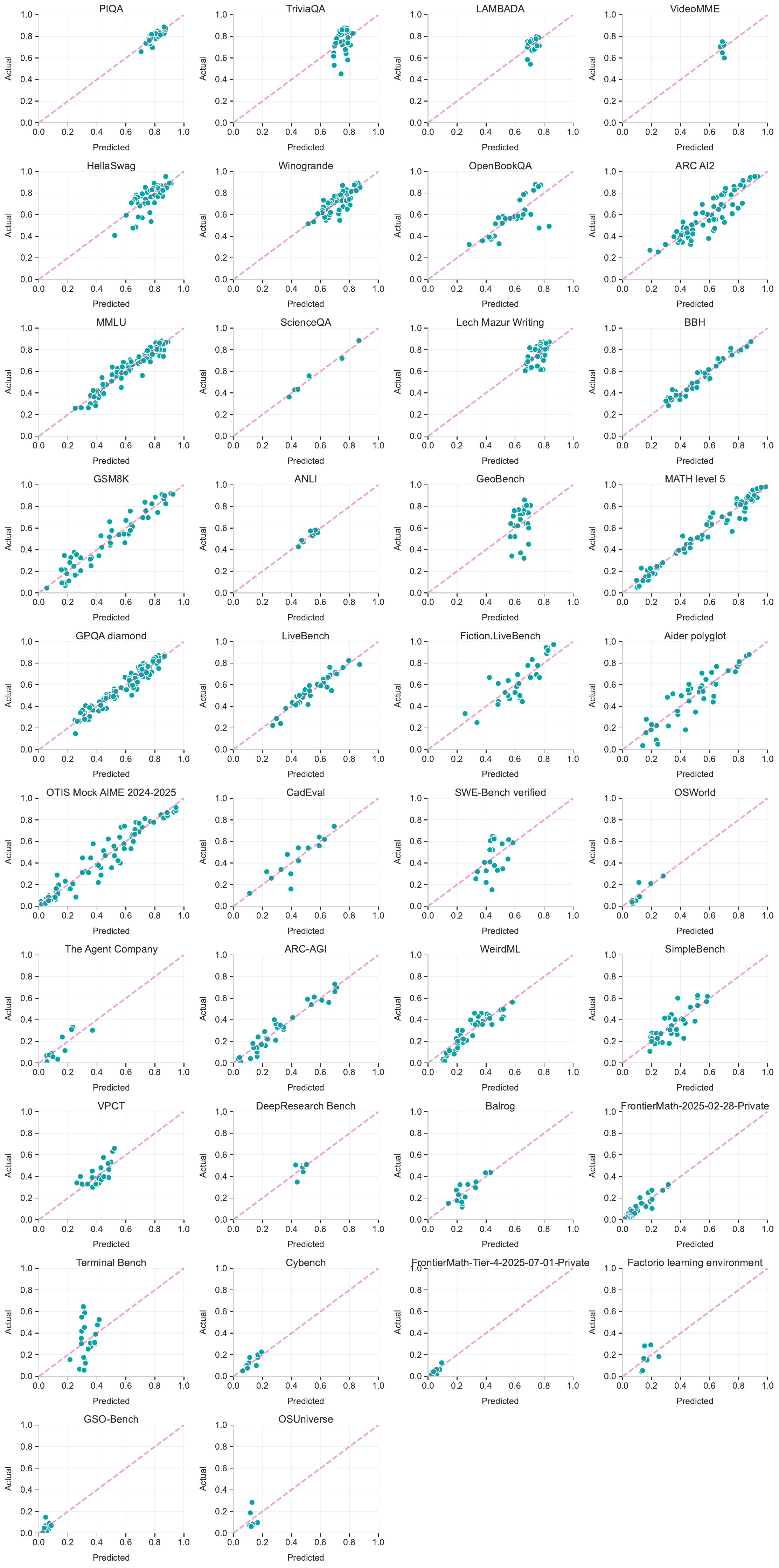}
    \caption{Our statistical framework predicts model scores fairly accurately.}
    \label{fig:predicted_vs_actual}
\end{figure}

In order to visualize how well our statistical framework does or does not fit the benchmark performance data, we compare the framework's predictions to the actual observed scores on each benchmark. This is shown in Figure \ref{fig:predicted_vs_actual}. We see that our statistical framework fits the data fairly accurately on most benchmarks, though there are a handful of exceptions like GeoBench where this is not the case. There are also benchmarks like CommonSenseQA 2.0 where we have little data to work with, and benchmarks like Cybench where most of the data points are clustered at low scores. 

\end{document}